%% file: main.tex
\documentclass[10pt,twocolumn,letterpaper]{article}
\usepackage{multirow}

\usepackage[pagenumbers]{cvpr} 
\usepackage{colortbl}
\usepackage{xcolor}

\definecolor{cvprblue}{rgb}{0.21,0.49,0.74}
\usepackage[pagebackref,breaklinks,colorlinks,allcolors=cvprblue]{hyperref}
\usepackage[ruled, lined, linesnumbered, commentsnumbered, longend]{algorithm2e}
\usepackage{mathtools}

\title{SharpDepth: Sharpening Metric Depth Predictions Using Diffusion Distillation}

\author{Duc-Hai Pham$^{1*}$ \quad Tung Do$^{1*}$ \quad Phong Nguyen$^{1}$ \quad Binh-Son Hua$^{1,2}$ \quad Khoi Nguyen$^{1}$ \quad Rang Nguyen$^{1}$ \\
\textsuperscript{1}VinAI Research, Vietnam, \textsuperscript{2}Trinity College Dublin \\
\texttt{\small \{v.haipd13, v.tungdt33, v.phongnh31, v.khoindm, v.rangnhm\}@vinai.io} \\ 
\texttt{\small binhson.hua@tcd.ie}\\
}

\def\Approach{SharpDepth}
\usepackage{svg}
\usepackage{newfloat}

\input{definition}

\input{sec/define}
\begin{document}

\twocolumn[{
\maketitle
\begin{center}
    \captionsetup{type=figure}
    \includegraphics[width=\textwidth]{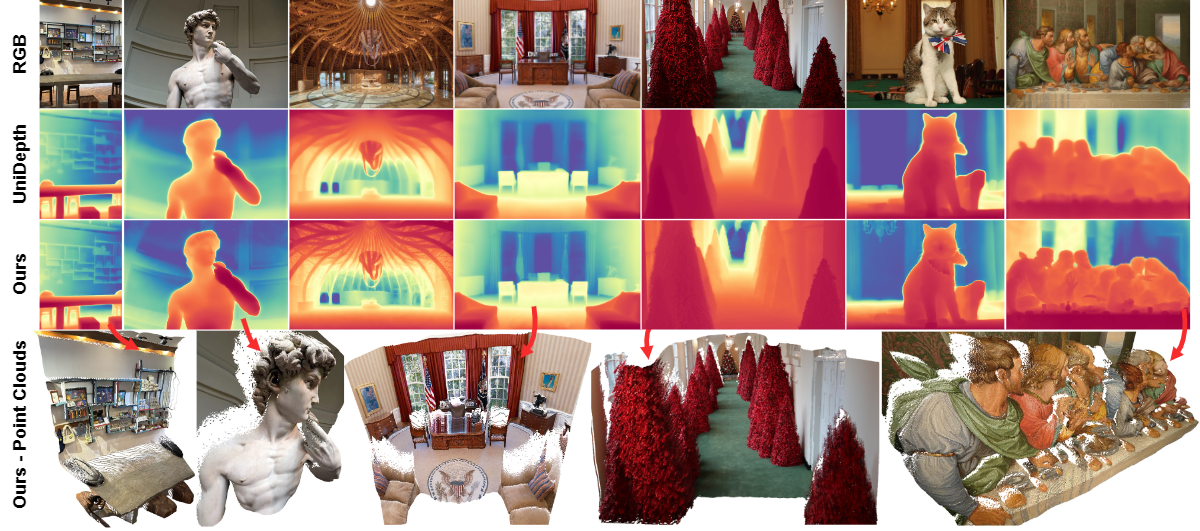}
    \captionof{figure}{We present SharpDepth, a diffusion-based depth model for refining metric depth estimators, e.g., UniDepth~\cite{piccinelli2024unidepth}, without relying on ground-truth depth data. Our method can recover sharp details in thin structures and improve overall point cloud quality.}
    \label{fig:teaser}
\end{center}
}]

\input{sec/0_abstract}    
\input{sec/1_intro}

\input{sec/2_related_work}

\input{sec/3_preliminaries}

\input{sec/4_methodology}

\input{sec/experiments}

\input{sec/5_conclusion}

\clearpage

\maketitlesupplementary

\input{sec/X_suppl_arxiv}

\clearpage

{
    \small
    \bibliographystyle{ieeenat_fullname}
    \bibliography{main}
}


\end{document}

%% file: definition.tex
\def\mI{\mathcal{I}}

\def\mL{\mathcal{L}}

\def\mN{\mathcal{N}}

\def\1n{\mathbf{1}_n}
\def\0{\mathbf{0}}
\def\1{\mathbf{1}}

\def\G{{\bf G}}

\def\bepsilon{\mbox{\boldmath{$\epsilon$}}}

\def\bdelta{\mbox{\boldmath{$\delta$}}}

\newcommand{\cm}[1]{}

\newcommand{\myheading}[1]{\vspace{1ex}\noindent \textbf{#1}}

\newif\ifshowsolution
\showsolutiontrue
\ifshowsolution

\else

\fi

\newcommand{\Eref}[1]{Eq.~(\ref{#1})}

%% file: sec/define.tex
\newcommand{\img}{\mI}

\newcommand{\latent}{z_{i}}

\newcommand{\discridepth}{d}
\newcommand{\genedepth}{\tilde{d}}

\newcommand{\blendedlatent}{z^\prime_{d}}
\newcommand{\metriclatent}{z_{d}}

\newcommand{\predlatentdepth}{\hat{z}}

\newcommand{\preddepth}{\hat{d}}

\newcommand{\encoder}{\mathcal{E}}

\newcommand{\discri}{f_D}
\newcommand{\genera}{f_G}

\newcommand{\refiner}{\G_\theta}
\newcommand{\refinerema}
{\G_{\bar{\theta}}}

\definecolor{tabfirst}{rgb}{1, 0.7, 0.7}
\definecolor{tabsecond}{rgb}{1, 0.85, 0.7}
\definecolor{tabthird}{rgb}{1, 1, 0.7}

%% file: sec/0_abstract.tex
\begin{abstract}
We propose \Approach, a novel approach to monocular metric depth estimation that combines the metric accuracy of discriminative depth estimation methods (e.g., Metric3D, UniDepth) with the fine-grained boundary sharpness typically achieved by generative methods (e.g., Marigold, Lotus). Traditional discriminative models trained on real-world data with sparse ground-truth depth can accurately predict metric depth but often produce over-smoothed or low-detail depth maps. Generative models, in contrast, are trained on synthetic data with dense ground truth, generating depth maps with sharp boundaries yet only providing relative depth with low accuracy. Our approach bridges these limitations by integrating metric accuracy with detailed boundary preservation, resulting in depth predictions that are both metrically precise and visually sharp. 
Our extensive zero-shot evaluations on standard depth estimation benchmarks confirm SharpDepth’s effectiveness, showing its ability to achieve both high depth accuracy and detailed representation, making it well-suited for applications requiring high-quality depth perception across diverse, real-world environments.
\end{abstract}

%% file: sec/1_intro.tex
\section{Introduction}
\label{sec:intro}

\begin{figure}[t]
  \centering
  \includegraphics[width=0.9\linewidth]{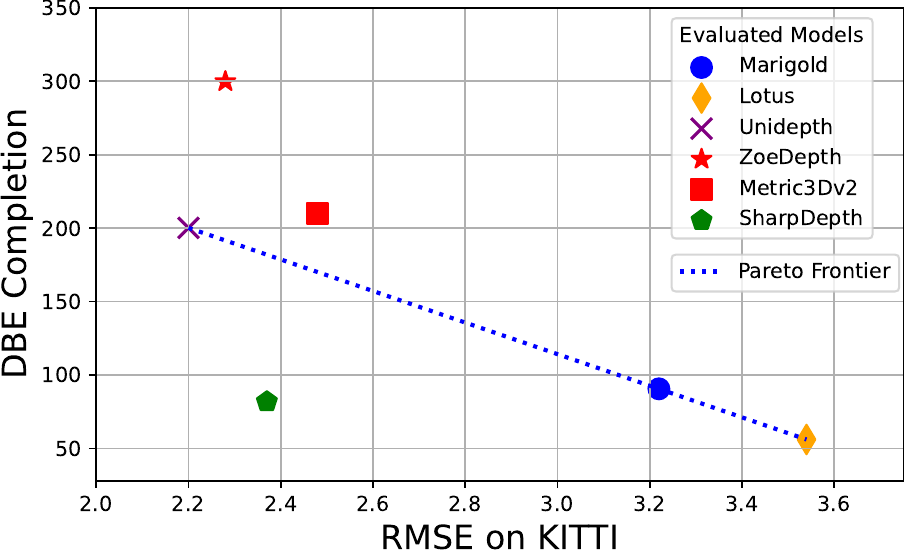}
  \caption{The performance of SOTA depth estimation models in terms of depth accuracy (x-axis) on KITTI \cite{baruch2021arkitscenes} and DBE Completion (y-axis) on Sintel\cite{sintel}, UnrealStereo4K \cite{unrealstereo} and Spring \cite{mehl2023spring}. Our method (SharpDepth) is best balanced on both axes.}
    \vspace{-10pt}
  \label{fig:pareto}
\end{figure}

Monocular metric depth estimation -- the task of predicting absolute depth from a single RGB image -- has emerged as a key computer vision problem due to its broad applications in autonomous driving~\cite{godard2019digging, guizilini20203d, guizilini2022full}, augmented reality~\cite{watson2019self, yucel2021real}, robotics~\cite{dong2022towards}, and 3D scene understanding~\cite{choi2015indoor, wang2021can}. Unlike stereo or multi-view approaches that leverage multiple viewpoints to deduce depth, monocular depth estimation aims to infer depth from a single perspective. This approach offers edges in terms of cost, hardware simplicity, and deployment flexibility; however, it also faces significant challenges due to inherent scale ambiguity and the limited depth cues present in single images. The challenge becomes more intense in the zero-shot setting, where no fine-tuning is performed on the target domain.

In the current literature on zero-shot monocular depth estimation, approaches generally fall into two categories: discriminative and generative methods. 
The discriminative approach~\cite{piccinelli2024unidepth,hu2024metric3d,zoedepth} typically relies on supervised learning with real-world data annotated by sparse ground truth depth, such as LiDAR measurements. These models are trained to regress metric depth values directly, and they generally provide accurate global depth estimates, capturing the true scale of the scene. However, due to the sparsity of ground-truth data and the reliance on coarse depth cues, these models often struggle with fine details, producing depth maps that tend to be blurry or lacking in edge sharpness, especially around object boundaries.

Recently, several generative depth estimation models have leveraged diffusion-based text-to-image models, excelling at producing depth maps with high spatial detail and sharp object edges~\cite{marigold,depth_anything_v1}. This improvement is attributed to the rich image priors inherited from large-scale text-to-image models. 
However, due to limitations inherent in latent diffusion models, fine-tuning these image-conditioned depth models is feasible only with synthetic data where dense depth maps are available, leading to a domain gap when applied to real images. 
Additionally, these models provide only affine-invariant depth rather than accurate metric depth. This limitation restricts their applicability in scenarios where precise metric depth information is essential. Some models can be modified to produce metric depth~\cite{depth_anything_v1}, which  requires additional metric information for training.

In this work, we introduce SharpDepth, a diffusion-based model designed to generate accurate zero-shot metric depth with high-frequency details. 
Built upon an affine-invariant depth estimator~\cite{marigold,he2024lotus}, our method refines the initial predictions of an existing metric depth model, enhancing the depth details while retaining accurate metric predictions. 
To this end, we measure the agreement of the depth predictions by the affine-invariant model and the metric depth model by normalizing both depth predictions to a common scale and produce a \emph{difference map}. 
Such a difference map allows us to identify image regions with reliable depth predictions as well as inconsistent depth regions where further sharpening and refinement are required. 

Based on the difference map, we propose to exploit the strengths of both the discriminative and generative depth estimators as follows. We propose Noise-aware Gating, a mechanism that guides the depth diffusion model to focus more precisely on regions identified as uncertain in the difference map. To further ensure both sharpness and accurate metric depth in these uncertain regions, we utilize two loss functions. 
We first leverage Score Distillation Sampling (SDS) loss to enhance depth detail, resulting in output with sharpness comparable to that of diffusion-based depth estimation methods. 
We then apply a Noise-aware Reconstruction loss to recognize the lack of scale awareness of diffusion-based models. This loss acts as a regularizer, ensuring that the final predictions remain close to the initial depth estimates, maintaining metric accuracy without drifting from the original depth scale. Together, these techniques enable SharpDepth to deliver precise, high-detail metric depth estimations across diverse scenes. Another benefit of the above training losses is that we can train our refinement on real data using only pretrained depth models, \textit{without} any additional ground truth for supervision.

To evaluate the performance of SharpDepth, we conduct extensive comparisons between our method with the state-of-the-art methods on both discriminative~\cite{piccinelli2024unidepth,zoedepth,hu2024metric3d} and generative methods~\cite{he2024lotus, marigold}. Experimental results show that our method 
can achieve both accurate in-depth estimation and still preserve high degrees of sharpness compared to the state-of-the-art metric depth estimators, e.g., UniDepth~\cite{piccinelli2024unidepth}, as shown in \cref{fig:teaser} and \cref{fig:pareto}.

In summary, our contributions are as follows:
\begin{itemize}
    \item We introduce \Approach, a novel diffusion-based depth sharpener model that can produce zero-shot metric depth with high-fidelity details.
    \item Our method can be trained with only images thanks to our two noise-aware proposed modules. The total amount of training images is about 100-150 times smaller than existing monocular depth estimation methods.
    \item Experiments on various zero-shot datasets show that our method's accuracy is competitive with discriminative models while containing the high-detail output of generative models.
\end{itemize}

%% file: sec/2_related_work.tex
\section{Related Work}

Current approaches to zero-shot monocular depth estimation can be divided into two main research lines: metric depth methods and affine-invariant depth methods.

\myheading{Zero-shot metric monocular depth estimation.}
Recently, a new line of metric monocular depth estimation (MMDE) methods has emerged, aiming to predict metric depth, also known as \textit{absolute depth}, without images from the target domain. For example, ZoeDepth \cite{zoedepth} achieves metrically accurate zero-shot predictions by fine-tuning a scale-invariant model on combined indoor and outdoor datasets, employing adaptive domain-specific range predictions. Concurrently, ZeroDepth \cite{zerodepth} introduces a transformer-based encoder-decoder that leverages camera intrinsics to enhance camera awareness, allowing it to directly decode metric depth without adaptive range prediction. Metric3D \cite{yin2023metric3d} and its enhanced version, Metric3Dv2 \cite{hu2024metric3d}, achieve zero-shot single-view metric depth through large-scale training and address metric ambiguity across camera models by incorporating a canonical camera space transformation module. UniDepth \cite{piccinelli2024unidepth} further advances MMDE by eliminating the need for test-time camera intrinsics, instead using a pseudo-spherical representation and a self-promoting camera module to predict both depth and camera parameters from a single image. However, these MMDE methods often yield depth predictions with limited detail due to the sparse ground truth in real-world data. In contrast, our approach delivers both high accuracy and fine detail, particularly in boundary regions.

\myheading{Zero-shot affine-invariant monocular depth estimation.}
Monocular depth estimation is an ill-posed geometric problem, and many approaches~\cite{eigen2014depth, lee2019big, fu2018deep, dpt, chen2020oasis, eftekhar2021omnidata} address this by estimating depth up to an unknown global scale and shift, also known as \textit{affine-invariant depth}. MegaDepth \cite{MegaDepthLi18} and DiverseDepth \cite{yin2021virtual} leverage large-scale internet images to achieve zero-shot depth estimation, though their training data often includes noisy labels. MiDAS \cite{Ranftl2022} mitigates this issue by using 3D movie frames with scale-shift-invariant losses to ensure consistency across various depth representations. Depth Anything \cite{depth_anything_v1} builds on this approach, employing a pseudo-labeling strategy across 62 million unlabeled images to enhance performance in real-world scenarios.
Recently, diffusion models have shown promise in improving depth fidelity by incorporating image priors. Marigold \cite{marigold} fine-tunes the Stable Diffusion model to generate high-quality depth with clear boundaries, while Lotus \cite{he2024lotus} accelerates inference by optimizing the noise scheduling process. GeoWizard \cite{fu2025geowizard} jointly estimates depth and normals, leveraging cross-modal relationships. Despite these advancements, the limitations of synthetic data create domain gaps that hinder the performance of diffusion-based methods in real-world applications, where discriminative feed-forward models still outperform them in terms of accuracy. Furthermore, these methods typically provide relative depth (i.e., depth relationships within a scene) rather than precise metric depth, which restricts their applicability in scenarios requiring accurate metric depth information.

\myheading{Affine-invariant and metric depth refinement.}
Recent work has focused on depth refinement rather than training models from scratch to leverage the benefits of both affine-invariant and metric methods. BetterDepth \cite{zhang2024betterdepth} refines affine-invariant depth by conditioning on outputs from pre-trained models and applying a generative model with a diffusion loss. PatchRefiner \cite{li2025patchrefiner} builds on this approach for metric refinement by using features from a metric base network to generate residual depth maps, which enhances detail in the final prediction. However, both methods rely heavily on synthetic datasets, limiting their applicability in real-world scenarios. In contrast, our method employs a ground-truth-free fine-tuning protocol, utilizing real-world data without annotations. This reduces reliance on synthetic datasets, minimizes domain gaps, and improves fine detail while preserving metric accuracy.

%% file: sec/3_preliminaries.tex
\section{Preliminaries}

\myheading{Diffusion Model for Depth Prediction.} 
To transfer the rich visual knowledge of diffusion models to the depth estimation domain, Marigold \cite{marigold} fine-tunes Stable Diffusion \cite{rombach2021highresolution} for monocular depth estimation. This process reuses the VAE encoder $\encoder$ on depth images to obtain depth latents $z_d$, optimizing with the $\epsilon$-prediction objective function in \cref{eq:marigold}. Additionally, $z_d$ is concatenated with the conditional image latent $z_i$, while the original text condition is omitted, $\epsilon \sim \mN(0,I)$, $z_d^t$ is the noisy version of $z_d$ at timestep $t$, and $\epsilon_\theta$ is the output of the diffusion model.
\begin{equation}
    \mathcal{L}_\theta = \mathbb{E}_{t, \epsilon \in \mathcal{N}(0,1)} \| \epsilon_\theta(z_d^t, t, z_i) - \epsilon \|_2^2.
\label{eq:marigold}
\end{equation}
Similarly, Lotus \cite{he2024lotus} is based on the same principles but introduces key modifications. It reduces the number of timesteps from 1000 to 1, enabling faster inference. Additionally, $z_0$-prediction is employed to reduce variance. To prevent catastrophic forgetting, Lotus jointly learns to predict both depth and image latents. The final training objective, as shown in \cref{diffusion:lotus}, defines $f_\theta$ as an $z_0$-prediction diffusion model, where $s_d$ and $s_i$ are task indicators for depth and image prediction, respectively.
%
\begin{align}
\mathcal{L}_\theta = \| z_d - f_\theta(z_d^t, z_i, t, s_d) \|_2^2 
+  \| z_i - f_\theta(z_i^t, t, s_i) \|_2^2.
\label{diffusion:lotus}
\end{align}

\myheading{Score Distillation Sampling (SDS)} is a distillation technique applied in 3D assets synthesis \cite{poole2022dreamfusion, sjc, lin2023magic3d}. By removing the U-Net Jacobian term of the gradient of  $\mL_{\text{diff}}$, a differentiable image $x=g(\phi)$ could be optimized without the need for backpropagating through the diffusion model U-Net. The gradient of SDS loss can be approximated as follows: 

\begin{equation}
    \nabla_\phi \mathcal{L}_{\text{SDS}}(\theta, \mathbf{x} = g(\phi)) \triangleq \mathbb{E}_{t, \epsilon} \bigg[ w(t) \big(\epsilon_\theta(z_i^t, y, t)  - \epsilon \big) \bigg],
\label{eq:sds}
\end{equation}
where $w(t)$ is the weighting at timestep $t$ and $y$ is the text prompt.

%% file: sec/4_methodology.tex
\begin{figure*}[t]
    \centering
    \includegraphics[width=0.9\textwidth]{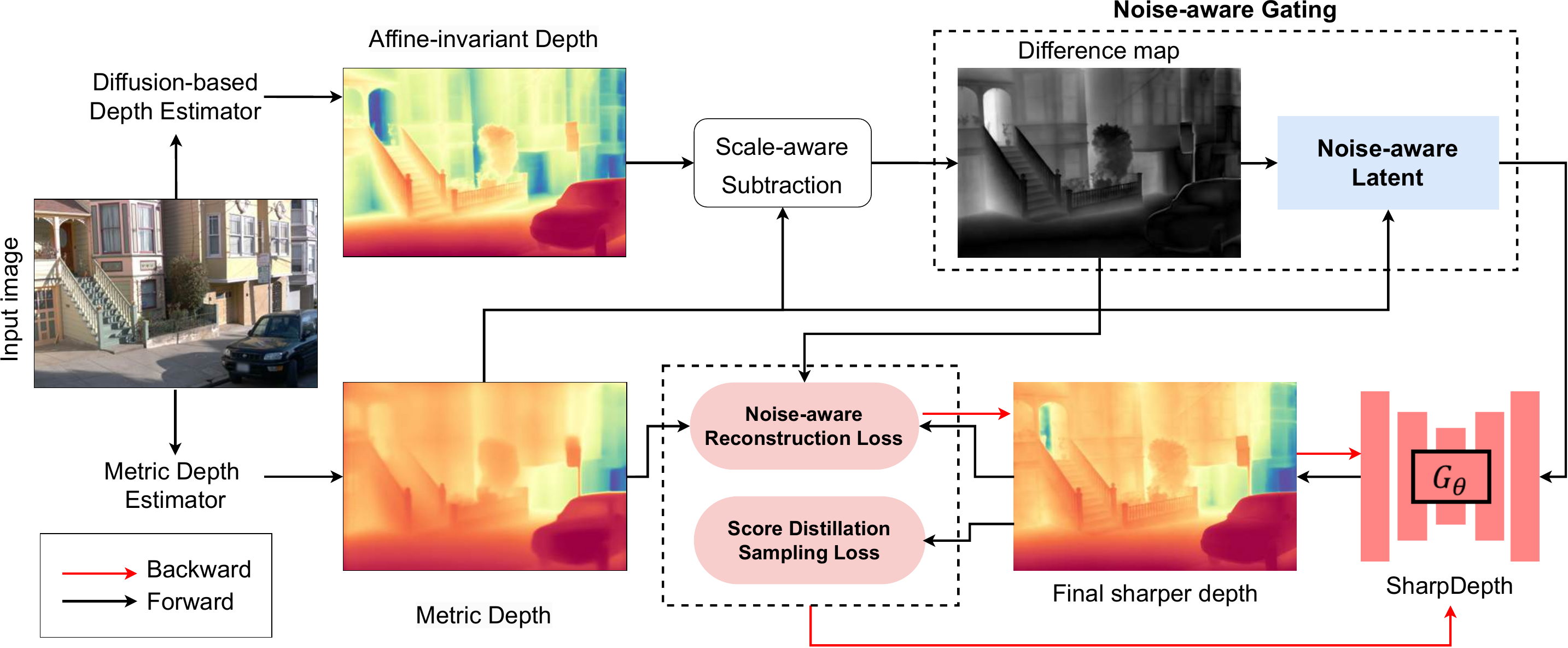}
    \caption{
    Our framework utilizes a diffusion-based estimator and a metric depth estimator to generate affine-invariant and metric depth maps, respectively. A Noise-Aware Gating mechanism produces a selectively noisy latent map, which is fed into our SharpDepth model. The training pipeline uses Score Distillation Sampling and Noise-Aware Reconstruction Losses to refine accuracy and enhance details.}
    \label{fig:method-train}
\end{figure*}

\begin{figure}[t]
  \centering
  \includegraphics[width=0.9\linewidth]{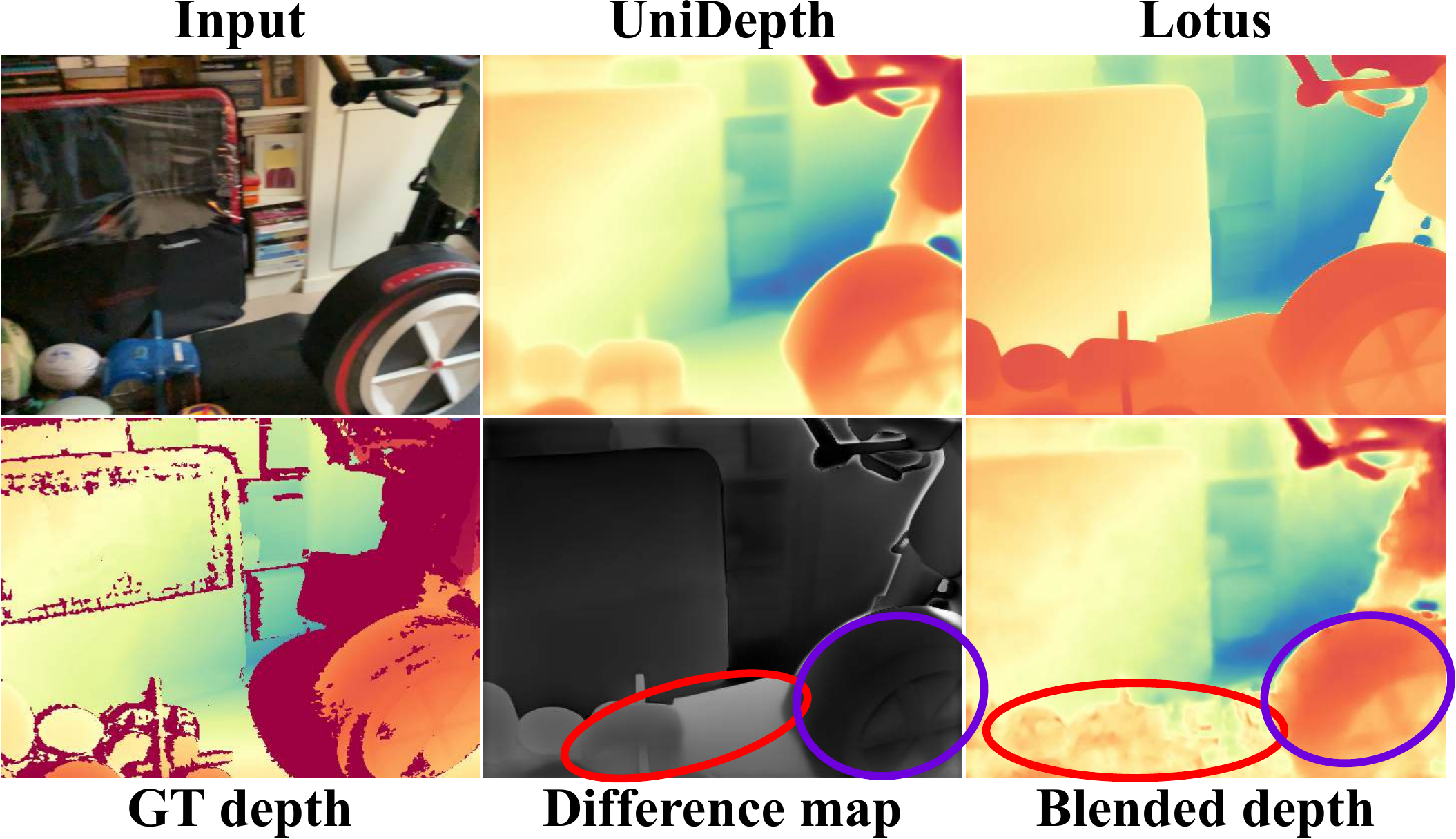}
  \caption{The difference map between the Unidepth and Lotus predictions. The \textcolor{red}{high-difference} (brighter) areas are heavily distorted by noise, whereas in the \textcolor{violet}{low-difference} (darker) areas, some information about the wheel is still recognizable.}
  \label{fig:example_noise}
  \vspace{-5pt}
\end{figure}

\section{Methodology}
Given an input image $\img$, we first use both 
pre-trained metric depth model $\discri$ and diffusion-based depth model $\genera$
to produce metric and affine-invariant depth output $\discridepth$ and $\genedepth$, respectively. 
Our goal is to generate a sharpened metric depth map, $\preddepth$, using our proposed sharpening model, $\refiner$. This model architecture is based on state-of-the-art pre-trained depth diffusion models \cite{marigold, he2024lotus}.

Instead of naively relying on the forward process of diffusion model \cite{ho2020denoising, rombach2021highresolution}, we introduce a Noise-aware Gating mechanism (described in \cref{sec:mask}), which provides explicit guidance to the sharpener $\refiner$ on uncertain regions. To enable ground-truth free fine-tuning, we use SDS loss to distill fine-grained details from the pretrained diffusion depth model $\genera$ and Noise-Aware Reconstruction Loss to ensure accurate metric prediction (as detailed in \cref{sec:loss}). The overall architecture is outlined in \cref{fig:method-train}.

\subsection{Noise-aware Gating}
\label{sec:mask}

Our goal is to develop a mechanism that enables the network to identify regions requiring sharpening. Given the challenges of obtaining or annotating per-pixel ground truth (GT) metric depths while maintaining overall sharp detail, we adopt an amortized approach that combines insights from both the diffusion-based depth estimator and a metric estimator. Intuitively, we assume that regions with minimal differences between these models are more reliable, while areas with larger discrepancies require additional supervision.
Using two state-of-the-art depth estimators, UniDepth~\cite{piccinelli2024unidepth} and Lotus~\cite{he2024lotus}, we find that UniDepth provides fairly accurate depth predictions compared to sparse GT data, though it lacks the sharp detail of Lotus's output. As shown in Fig.~\ref{fig:example_noise}, we first normalize these predictions and then compute a difference map between two inferred depth maps. Brighter areas in the difference map highlight regions with significant discrepancies, indicating areas needing further refinement. Conversely, darker regions represent areas of mutual agreement between the depth estimators. While these regions may still benefit from depth refinement, they hold lower priority in the sharpening process.

To this end, we propose a Noise-aware Gating mechanism incorporating information from the initial metric depth $\discridepth$ as input to our $\refiner$ model. Advances in image inpainting~\cite{ju2024brushnet}, editing~\cite{repaint}, and virtual try-on~\cite{tryondiffusion} have used explicit masks to guide diffusion models to focus on regions of interest. Inspired by these models, we avoid adding pure Gaussian noise to every pixel of the clean latent depth. Instead, we selectively introduce noise to regions with significant differences, while applying less noise to areas with smaller discrepancies. This strategy effectively directs the sharpener to focus on high-difference (noisy) regions, leaving low-difference (clean) areas mostly unaffected.

To align the depth ranges of 
$\genedepth$ and $\discridepth$, we first scale and shift $\genedepth$ to the range of $\discridepth$ before calculating the difference map. The difference map $e$ is then computed as the absolute difference between the adjusted $\genedepth$ and $\discridepth$. 
As training progresses, we observe that our proposed 
$\refiner$ begins to generate depth maps superior to those produced by the diffusion-based model $\genera$. As the results, we replace $\genera$ with the exponential moving average (EMA) of the training model, 
$\refinerema$, which serves as a refined initialization for $\genedepth$ and enables iterative refinement in multiple steps.
Details regarding the performance comparison of these two approaches are provided in \cref{sec:ablate}.

Once the difference map 
$e$ is obtained, we use it to control the noise intensity applied to each region of $\metriclatent$, which is the latent representation of $\discridepth$. Specifically, we perform a weighted blending between Gaussian noise $\epsilon$ and 
$\metriclatent$ as follows: 
\begin{equation}
    \blendedlatent = \hat{e}\odot\epsilon + (1-\hat{e})\odot\metriclatent,
    \label{eq:blending}
\end{equation}
where $\hat{e}$ is the downsampled version of the difference map $e$ to match the dimensions of latent $\metriclatent$, and $\odot$ is the element-wise product.

This blended latent $\blendedlatent$ effectively distinguishes high- and low-difference regions between the two depth predictions, serving as a powerful prior for the sharpener as shown in \cref{fig:example_noise}. 
By separating these regions, the optimization process focuses on high-difference areas while minimizing modifications in similar regions, enabling the sharpener to reconstruct fine-grained details while maintaining accuracy.

\subsection{Training Objectives}
\label{sec:loss}

\myheading{Diffusion Depth Prior Distillation.} In this section, we introduce our approach to distill the knowledge of $\genera$ into our depth sharpener, $\refiner$. Inspired by SwiftBrush \cite{nguyen2024swiftbrush} and DreamFusion \cite{poole2022dreamfusion}, we do not train our $\refiner$ from scratch, instead, perform score distillation on a pretrained diffusion-based depth estimator such as $\genera$. 
The output predicted latent $\predlatentdepth$ of our proposed model is computed as follows:  $\predlatentdepth = \refiner(\blendedlatent, \latent)$.
We then slightly modify the original SDS formulation of  $\epsilon$-prediction in the \cref{eq:sds} to match the $x_0$-prediction of $\genera$. The revised version of SDS loss for training our $\refiner$ can be defined as follows: 
\begin{equation} 
    \nabla_\theta \mathcal{L}_{\text{SDS}} \triangleq \mathbb{E}_{t, \epsilon} \bigg[ w^t \bigg(\predlatentdepth -\genera(\predlatentdepth^t; \latent, t) \bigg) \bigg],
\label{eq:sds_depth}
\end{equation}
where $\predlatentdepth^t$ is the noisy version of $\predlatentdepth$ at time step $t$.

\myheading{Noise-aware Reconstruction Loss.} Our distillation objective encourages the output of $\refiner$ to align more closely with the distribution of the diffusion model $\genera$, leading to highly detailed depth images. However, this also causes the network to inherit the limitations of $\genera$, ultimately reducing accuracy in metric depth estimation. To address this issue, we introduce an additional reconstruction loss that preserves the accuracy of the discriminative model by measuring the distance between our network’s output and the discriminative output in \cref{eq:l1_depth}.

Specifically, we use the difference map to enforce larger gradients in regions where $\discridepth$ and $\preddepth$ exhibit significant differences. This serves as an explicit regularization mechanism, encouraging the sharpener to focus more on these pixels. While the difference map ensures that regions with minimal differences remain largely unchanged, it may also risk propagating over-smoothing artifacts to $\preddepth$. The reconstruction loss is given by:
\begin{equation}
    \mathcal{L}_{\text{recons}} = \|e\odot(\preddepth - \discridepth)\|, \label{eq:l1_depth}
\end{equation} 
where $\odot$ is the element-wise multiplication.

\myheading{Discussion.} 
Though the two objective functions described above serve different purposes -- one enhancing depth detail and the other ensuring accurate depth values -- both focus on high-difference regions. High-difference regions are augmented with noise and subsequently refined by $\refiner$, resulting in large gradients during optimization. This causes Score Distillation Sampling (SDS) to emphasize these regions which function as an implicit masking optimization. In contrast, the reconstruction loss employs a difference map directly during optimization, resembling an explicit masking optimization approach.
The complete training objective function in \Eref{eq:loss} comprises $\mathcal{L}_{\text{SDS}}$ and $\mathcal{L}_{\text{recons}}$ where $\lambda_{\text{SDS}}$ and $\lambda_{\text{recons}}$ are loss weighting hyperparameters.
\begin{equation}
    \mL_{\text{total}} = \lambda_{\text{SDS}} \ \mathcal{L}_{\text{SDS}} + \lambda_{\text{recons}} \ \mathcal{L}_{\text{recons}}.
    \label{eq:loss}
\end{equation}

%% file: sec/experiments.tex
\section{Experiments}

\subsection{Experimental Setup}

\myheading{Datasets:} For training, we use approximately 1\% of each of the following real-world datasets, which cover various camera types and scene domains: Pandaset \cite{xiao2021pandaset}, Waymo \cite{sun2020scalability}, ArgoVerse2 \cite{wilson2023argoverse}, ARKit \cite{baruch2021arkitscenes}, Taskonomy \cite{zamir2018taskonomy}, and ScanNetv2 \cite{dai2017scannet}. This results in a combined training set of 90,000 images which is 100-150 times smaller than the amount of data has been used for discriminative depth models~\cite{yin2023metric3d,piccinelli2024unidepth}.


For testing, we evaluate our approach on seven real datasets:  KITTI \cite{geiger2013vision, eigen2014depth}, NYUv2 \cite{silberman2012indoor}, ETH3D~\cite{schops2017multi}, Diode \cite{vasiljevic2019diode}, Booster \cite{ramirez2023booster}, nuScenes \cite{caesar2020nuscenes}, and iBims~\cite{ibims} and three synthetic datasets: Sintel~\cite{sintel}, UnrealStereo4K~\cite{unrealstereo}, and Spring~\cite{mehl2023spring}.



\myheading{Metrics:} 
We use common depth estimation metrics: absolute mean relative error (A.Rel), root mean square error (RMSE), and the percentage of inlier pixels ($\mathrm{\delta}_{1}$) with a threshold of 1.25. Additionally, to assess the sharpness of predictions, we adopt the Depth Boundary Error (DBE) inspired by iBims~\cite{ibims}. Since annotated depth edges are not available for synthetic datasets, we adapt the original metric into a Pseudo Depth Boundary Error (PDBE). In PDBE, we apply Canny Edge Detection to the predicted and ground truth depths to generate sets of edges, which are then used to compute the accuracy $\epsilon^{\text{acc}}_{\text{PDBE}}$ and completion $\epsilon^{\text{compl}}_{\text{PDBE}}$, following the original iBims formulation. Further details are provided in the supplementary material.


\myheading{Implementation details.} 
We implement our method in PyTorch, using Lotus \cite{he2024lotus} and UniDepth \cite{piccinelli2024unidepth} as the diffusion-based and metric depth estimators, respectively. We also adopt the architecture and pretrained weights from Lotus for our SharpDepth model’s initialization. The Adam optimizer is used with a learning rate of $1e^{-6}$, and we set the loss weights as $\lambda_{\text{SDS}}=1.0$ and $\lambda_{\text{recons}}=0.3$. Optimization is performed for 13,000 iterations with a batch size of 1, using 16 gradient accumulation steps. Training our model to convergence takes approximately 1.5 days on two A100 40GB GPUs.
We normalize the metric depth from $\discri$ to $[-1, 1]$ before feeding it into the VAE encoder and revert the normalization after decoding. Min-max normalization is applied to the input, and the output is rescaled using least-squares alignment with the original metric depth. The difference map is applied to ensure alignment only for pixels with minimal differences.

\subsection{Comparison with the State of the Art}
\label{sec:compare_sota}

\myheading{Baselines.} We compare our method against four zero-shot metric depth models: UniDepth \cite{piccinelli2024unidepth}, Metric3Dv2 \cite{hu2024metric3d}, ZoeDepth \cite{zoedepth}, and PatchRefiner \cite{li2025patchrefiner}. Ground truth intrinsics are provided to UniDepth and Metric3Dv2. Since ZoeDepth is fine-tuned on KITTI and NYUv2, it does not qualify as a zero-shot model and is excluded from zero-shot evaluations. We also introduce a straightforward baseline, termed UniDepth-aligned Lotus, where we convert Lotus to metric depth by simply applying scale and shift adjustments based on the aligned UniDepth.
Additionally, we include three zero-shot relative depth models: Marigold \cite{marigold}, Lotus \cite{he2024lotus}, and BetterDepth \cite{zhang2024betterdepth} for reference purposes since they require ground truth alignment during test time. 
%

\myheading{Quantitative results.}
As shown in \cref{tab:results:zeroshot:metric}, our method achieves accuracy on par with metric depth models, demonstrating the effectiveness of our training pipeline. This approach enables the sharpener to selectively refine uncertain regions while preserving confident areas intact. Although PatchRefiner is trained on a comparable dataset size to \Approach, its performance is robust primarily on outdoor datasets. By leveraging our ground-truth-free fine-tuning pipeline, we are able to train across diverse real-world datasets, enabling strong generalization to both indoor and outdoor scenes.
Furthermore, our approach consistently surpass the naive UniDepth-aligned Lotus baseline across all datasets, demonstrating that relying solely on UniDepth's output for alignment is suboptimal. Our approach addresses this by using a difference map as a powerful guide, enabling alignment only on reliable pixels.

Also, \cref{tab:results:zeroshot:detail} provides the evaluation of depth details on 3 synthetic (Sintel~\cite{sintel}, UnrealStereo~\cite{unrealstereo} and Spring~\cite{mehl2023spring}) and 1 real dataset (iBims~\cite{ibims}). Compared to UniDepth, \Approach~obtains significantly higher results in both edge accuracy and completion. By leveraging rich priors from the pre-trained diffusion model, our method can produce sharper depth discontinuities, leading to high accuracy scores on all datasets. On the other hand, discriminative-based methods frequently produce smooth edges without clear transitions between objects. This is highlighted in the completeness error, as missing edges are expressed by large values of this error. 
While UniDepth-aligned Lotus depth details are competitive, high-frequency details that lack accurate metric precision on real datasets significantly limit the model's application.

\begin{figure}[t]
  \centering
  \includegraphics[width=\linewidth]{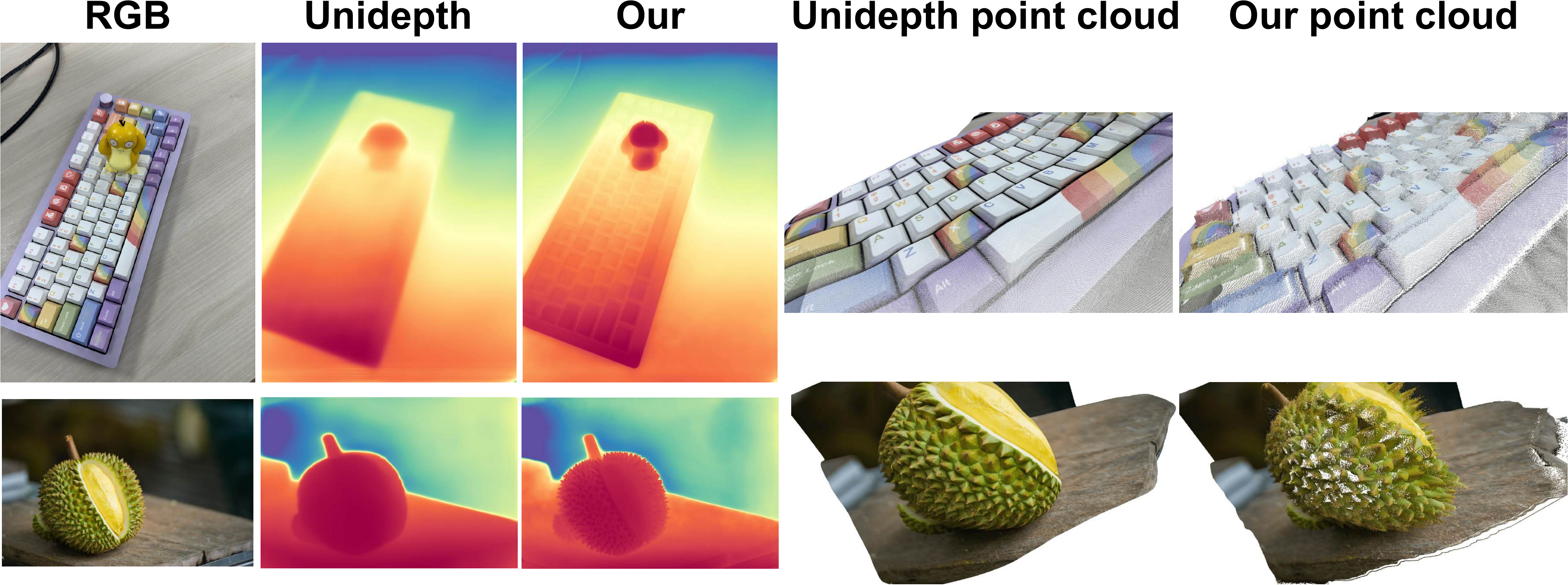}
  \caption{\textbf{Un-projected point cloud from in-the-wild image.}}
  \label{fig:exp_pcl}
\vspace{-5pt}
\end{figure}

\myheading{Qualitative Results.}
For a visual assessment, we present qualitative results in \cref{fig:results:main_vis} and \cref{fig:exp_pcl}. As shown in \cref{fig:results:main_vis}, \Approach~encompassed both high-frequency details in thin structures (fences and traffic poles in the first example) and accurate metric scene layout (second row). Moreover, detailed depth maps allow for better object reconstruction. To show this, we un-project the predicted depth maps to point-clouds from both \Approach~and UniDepth in~\cref{fig:exp_pcl}. \Approach~demonstrates better reconstruction fidelity, accurately capturing intricate details like the spikes on a durian and the contours of keyboard keycaps.


\begin{table*}
    \centering
    \resizebox{\linewidth}{!}{
    \begin{tabular}{lc cc cc cc cc cc cc}
        \toprule
        \multirow{2}{*}{\textbf{Method}} & \textbf{GT}&\multicolumn{2}{c}{\bf KITTI} & \multicolumn{2}{c}{\bf NYUv2} & \multicolumn{2}{c}{\bf ETH3D} & \multicolumn{2}{c}{\bf Diode} & \multicolumn{2}{c}{\bf Booster} & \multicolumn{2}{c}{\bf NuScenes}   \\
        \cmidrule(lr){3-4} 
        \cmidrule(lr){5-6}
        \cmidrule(lr){7-8} 
        \cmidrule(lr){9-10}
        \cmidrule(lr){11-12}
        \cmidrule(lr){13-14}
         &\textbf{aligned?}& $\mathrm{\bdelta}_{1}\uparrow$ & $\textbf{A.Rel}\downarrow$  & $\mathrm{\bdelta}_{1}\uparrow$ & $\textbf{A.Rel}\downarrow$ & $\mathrm{\bdelta}_{1}\uparrow$ & $\textbf{A.Rel}\downarrow$ & $\mathrm{\bdelta}_{1}\uparrow$ & $\textbf{A.Rel}\downarrow$ & $\mathrm{\bdelta}_{1}\uparrow$ & $\textbf{A.Rel}\downarrow$ & 
         $\mathrm{\bdelta}_{1}\uparrow$ & $\textbf{A.Rel}\downarrow$   \\
        \toprule
        Marigold \cite{marigold} & \checkmark & 0.92 & 0.09 & 0.96 & 0.05 & 0.96 & 0.06 & 0.77 & 0.31 & 0.97 & 0.05 & 0.66 & 0.27\\
        Lotus \cite{he2024lotus} & \checkmark & 0.88 & 0.11 & 0.97 & 0.05  & 0.96 & 0.06 & 0.74 & 0.33 & 0.99 & 0.04 & 0.51 & 0.36 \\
        BetterDepth \cite{zhang2024betterdepth} & \checkmark & 0.95 & 0.75 & 0.98 & 0.04 & 0.98 & 0.05 & - & - & - & - & - & - \\
        \midrule
         UniDepth \cite{piccinelli2024unidepth} &  & \cellcolor{tabfirst!170}{0.98} & \cellcolor{tabfirst!170}{0.05} & \cellcolor{tabfirst!170} 0.98 & \cellcolor{tabfirst!170} 0.05 & \cellcolor{tabthird!170} 0.25 & \cellcolor{tabsecond!170} 0.46 & \cellcolor{tabsecond!170} 0.66 & \cellcolor{tabsecond!170} 0.26 & \cellcolor{tabfirst!170} 0.28 & \cellcolor{tabfirst!170} 0.49 & \cellcolor{tabfirst!170} 0.84 & \cellcolor{tabfirst!170} 0.14 \\
        ZoeDepth \cite{zoedepth} & & \cellcolor{gray!50}{0.97} & \cellcolor{gray!50}{0.06}  & \cellcolor{gray!50}{0.95} & \cellcolor{gray!50}{0.08} & \cellcolor{tabsecond!170} 0.34 & 0.58  & 0.30 & 0.48  & \cellcolor{tabthird!170} 0.21 & \cellcolor{tabsecond!170} 0.64 & 0.22 & 0.59 \\
        Metric3Dv2 \cite{hu2024metric3d} & & \cellcolor{tabfirst!170}{0.98} & \cellcolor{tabfirst!170}{0.05} & \cellcolor{tabsecond!170}0.97 & \cellcolor{tabthird!170} 0.07 & \cellcolor{tabfirst!170} 0.82 & \cellcolor{tabfirst!170}0.14 & \cellcolor{tabfirst!170} 0.88 & \cellcolor{tabfirst!170} 0.16 & 0.15 & \cellcolor{tabthird!170}0.67 & \cellcolor{tabfirst!170} 0.84 & \cellcolor{tabthird!170} 0.20\\
        PatchRefiner \cite{li2025patchrefiner} & & 0.79 & 0.16 & 0.01 & 2.48 & 0.05 & 1.78 & 0.25 & 1.26 & 0.01 & 5.55 & 0.32 & 0.58 \\
        UniDepth-aligned Lotus & & \cellcolor{tabthird!170}0.84 & \cellcolor{tabthird!170}0.13 & \cellcolor{tabthird!170}0.94 & 0.09 & 0.20 & 0.49 & 0.56 & 0.36 & \cellcolor{tabsecond!170} 0.26 & \cellcolor{tabfirst!170} 0.49 & \cellcolor{tabthird!170} 0.41 & 0.43\\
        \midrule
        \Approach~(ours) &  &  \cellcolor{tabsecond!170}{0.97} & \cellcolor{tabsecond!170}{0.06} & \cellcolor{tabsecond!170} 0.97 & \cellcolor{tabsecond!170}0.06 & 0.23 & \cellcolor{tabthird!170} 0.47 & \cellcolor{tabthird!170} 0.61 & \cellcolor{tabthird!170} 0.29 & \cellcolor{tabfirst!170} 0.28 &\cellcolor{tabfirst!170} 0.49 & \cellcolor{tabsecond!170} 0.78 & \cellcolor{tabsecond!170} 0.18 \\

        \bottomrule
    \end{tabular}
    }
    \vspace{-3pt}
    \caption{\textbf{Comparison for depth accuracy purpose on real-image datasets.} These methods are trained and tested on non-overlapping datasets. Our SharpDepth results use the initial metric depth prediction of UniDepth. `-' indicates not reported results. GT-aligned indicates the method has to use GT depth to align in testing. We ranked methods that do not require GT alignment as \colorbox{tabfirst!170}{best}, \colorbox{tabsecond!170}{second-best}, and \colorbox{tabthird!170}{third-best}. \colorbox{gray!50}{Gray} indicates the method that has been trained on the training set.} 
    \label{tab:results:zeroshot:metric}
\end{table*}

\begin{table*}[t]
    \centering
    \resizebox{\linewidth}{!}{
    \begin{tabular}{l ccc ccc ccc ccc}
        \toprule
        \multirow{2}{*}{\textbf{Method}} & \multicolumn{3}{c}{\bf Sintel} & \multicolumn{3}{c}{\bf UnrealStereo4K} & \multicolumn{3}{c}{\bf Spring} & \multicolumn{3}{c}{\bf iBims} \\
         \cmidrule(lr){2-4} 
         \cmidrule(lr){5-7} 
         \cmidrule(lr){8-10} 
         \cmidrule(lr){11-13} 
         & $\bepsilon^{\textbf{acc}}_{\textbf{PDBE}}\downarrow$ & $\bepsilon^{\textbf{compl}}_{\textbf{PDBE}}\downarrow$ &
         $\textbf{A.Rel}\downarrow$  &
         $\bepsilon^{\textbf{acc}}_{\textbf{PDBE}}\downarrow$ & $\bepsilon^{\textbf{compl}}_{\textbf{PDBE}}\downarrow$ &
         \textbf{$\textbf{A.Rel}\downarrow$} &
        $\bepsilon^{\textbf{acc}}_{\textbf{PDBE}}\downarrow$ & $\bepsilon^{\textbf{compl}}_{\textbf{PDBE}}\downarrow$& 
        
         \textbf{$\textbf{A.Rel}\downarrow$ }& $\bepsilon^{\textbf{acc}}_{\textbf{DBE}}\downarrow$ & $\bepsilon^{\textbf{compl}}_{\textbf{DBE}}\downarrow$& 
         \textbf{$\textbf{A.Rel}\downarrow$} \\

        \toprule
        Marigold~\cite{marigold} & 1.90 & 52.5 & 0.65 & 1.97 & 68.6 &  0.56 & 1.85 & 150.3 & 0.69 & 1.85 & 13.4 & 0.07\\
        Lotus~\cite{he2024lotus} & 2.03 & 31.9 & 0.53 & 1.21 & 33.2 & 0.67 & 1.27 & 102.8 & 0.74 &1.92 & 11.0 & 0.07 \\
        \midrule
        UniDepth~\cite{piccinelli2024unidepth} & 3.73 & 113.3 & \cellcolor{tabthird!170} 0.96 & 8.65 & \cellcolor{tabthird!170} 257.3 & \cellcolor{tabsecond!170} 0.47 & 5.29 & 229.7 & \cellcolor{tabsecond!170} 0.66 & \cellcolor{tabthird!170} 2.00 & 30.0 & \cellcolor{tabthird!170} 0.38\\
        ZoeDepth~\cite{zoedepth} & 3.35 & \cellcolor{tabthird!170} 45.8 & 1.78 & 5.39 & 649.3 & 0.73 & 4.05 & 204.2 & 0.78 & 2.05 & 23.7 & \cellcolor{tabfirst!170} 0.17\\
        Metric3Dv2~\cite{hu2024metric3d} & \cellcolor{tabfirst!170} 1.92 & 63.8 & \cellcolor{tabfirst!170} 0.47 & \cellcolor{tabthird!170} 2.75 & 446.9 & \cellcolor{tabfirst!170} 0.38 & \cellcolor{tabthird!170} 1.78 & \cellcolor{tabsecond!170} 118.4 & \cellcolor{tabfirst!170} 0.56 & 2.14 & \cellcolor{tabsecond!170} 12.3 & \cellcolor{tabsecond!170} 0.19\\
        PatchRefiner~\cite{li2025patchrefiner} & 3.86 & 58.6 & 3.73 & 4.98 & 800.8 & 0.92 & 4.19 & 225.4 & 1.11 & 2.49 & 38.3 & 2.43\\
        UniDepth-aligned Lotus & \cellcolor{tabthird!170} 2.04 & \cellcolor{tabfirst!170} 31.9 &  0.97 & \cellcolor{tabfirst!170} 1.21 & \cellcolor{tabfirst!170} 33.2 & \cellcolor{tabthird!170} 0.53 & \cellcolor{tabsecond!170} 1.27 & \cellcolor{tabfirst!170} 102.7 & \cellcolor{tabthird!170} 0.74 & \cellcolor{tabsecond!170} 1.92 & \cellcolor{tabfirst!170} 11.0 & 0.39 \\
        \midrule
        \Approach~(ours) & \cellcolor{tabsecond!170} 1.94 &  \cellcolor{tabsecond!170} 36.2 & \cellcolor{tabsecond!170} 0.92 & \cellcolor{tabsecond!170} 1.37 & \cellcolor{tabsecond!170} 61.5 & \cellcolor{tabsecond!170} 0.47 & \cellcolor{tabfirst!170} 1.24 & \cellcolor{tabthird!170} 147.6 & \cellcolor{tabsecond!170} 0.66 & \cellcolor{tabfirst!170} 1.80 & \cellcolor{tabthird!170} 13.1 & 0.39\\
        \bottomrule
    \end{tabular}}
\vspace{-5pt}
\caption{\textbf{Comparison for depth details purpose on one real-image dataset (iBims) and three synthetic datasets (Sintel, UnrealStereo4K, Spring)}. We ranked methods that do not require GT alignment as \colorbox{tabfirst!170}{best}, \colorbox{tabsecond!170}{second-best}, and \colorbox{tabthird!170}{third-best}.} 
    
    \label{tab:results:zeroshot:detail}
\end{table*}

\begin{table}[t]
\footnotesize
\centering

\setlength{\tabcolsep}{0.18em}
\renewcommand{\arraystretch}{1.00}
\begin{tabular}{clccccc}
\toprule
\multirow{2}[2]{*}{\textbf{Setting}} &
\multirow{2}[2]{*}{\textbf{Method}} & 
\multicolumn{3}{c}{\textbf{KITTI}} &
\multicolumn{2}{c}{\textbf{Sintel}} 
\\
\cmidrule(lr){3-5} \cmidrule(lr){6-7}
& &  
$\mathrm{\delta}_{1}\uparrow$ & 
$\text{A.Rel}\downarrow$ & $\text{RMSE}\downarrow$&
$\epsilon^{\text{acc}}_{\text{DBE}}\downarrow$ & $\epsilon^{\text{compl}}_{\text{DBE}}\downarrow$ 
\\
\midrule
- & \textbf{Ours} & 0.973 & 0.060 & 2.37 & 1.94 & 36.4
\\
\midrule
A & Input Noise latent & 0.817 & 0.135 & 3.98 & 1.94 & 34.6
\\
B & Input $\metriclatent$ & 0.701 & 0.186 & 4.78 & 3.30 & 116.9

\\
\midrule
C & w/o SDS loss &  0.978 & 0.051 & 2.28 & 3.70 & 112.5
\\
D & w/o reconstruction loss & 0.843 & 0.128 & 3.66 & 1.94 & 34.1

\\
\midrule
E & Lotus teacher (ours) & 0.973 & 0.060 & 2.37 & 1.94 & 36.4
\\
F & Marigold teacher & 0.973 & 0.058 & 2.34 & 2.40 & 84.7 \\
\midrule
G & Frozen Lotus~\cite{he2024lotus} & 0.967 & 0.069 & 2.43 & 2.00 & 40.6 \\
H & EMA update (ours) & 0.973 & 0.060 & 2.37 & 1.94 & 36.4 \\

\bottomrule
\end{tabular}
\vspace{-3pt}
\caption{
\textbf{Ablation study} of different design choices. 
}
\label{tab:ablation}
\vspace{-5pt}

\end{table}

\input{sec/main_vis} 

\subsection{Ablation Study}
\label{sec:ablate}
In this section, we validate the impact of our design choices through experiments on two validation sets: KITTI \cite{baruch2021arkitscenes} and Sintel \cite{sintel}. Unless otherwise specified, all experiments use the same training dataset as described in \cref{sec:compare_sota}.

\myheading{Analysis of Noise-aware Gating} are presented in \cref{tab:ablation} (Settings A-B). Instead of calculating the difference map as in \cref{sec:mask} to generate input for $\refiner$, we test two alternative inputs: (A) Gaussian noise, as in \cite{he2024lotus}, and (B) the output of the pretrained metric depth model.
Both configurations led to decreased performance, with RMSE increasing by 1.64 in setting (A) and by 2.34 in setting (B). In setting (A), the lack of prior knowledge from the pretrained metric depth model led the sharpener to behave similarly to Lotus, degrading accuracy. In setting (B), although the pretrained metric depth estimator provides valuable prior knowledge, the absence of explicit guidance from the difference map $e$ results in ambiguity regarding which regions require refinement during reconstruction loss calculation. This makes it difficult for the model to balance the diffusion and metric depth priors, hindering training.

\myheading{Effects of the Training Objectives} discussed in \cref{sec:loss} are analyzed in \cref{tab:ablation} (Settings C-D). We remove each component to examine its impact on the sharpener. In setting (C), we exclude the SDS loss, while in setting (D), we remove the Noise-aware Reconstruction loss.
In setting (C), without the distillation loss to incorporate information from the pretrained diffusion-based depth estimator $\genera$, our model aligns more closely with $\discri$, producing nearly identical predictions. As a result, depth accuracy remains competitive, closely matching $\discri$, while setting (D) mirrors $\genera$, with high detail accuracy but lower depth accuracy.

\begin{figure}[t]
  \centering
  \includegraphics[width=\linewidth]{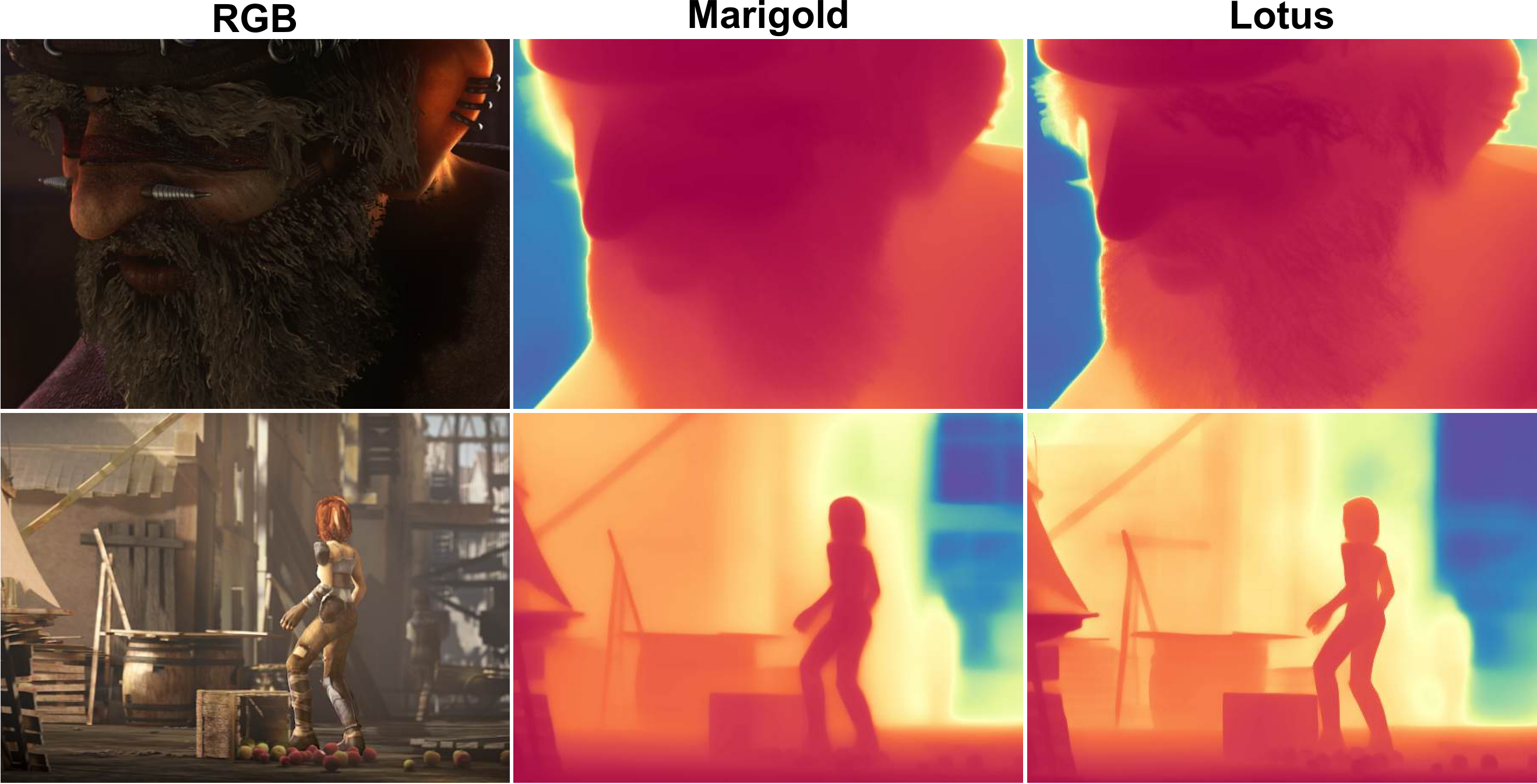}
  \caption{\textbf{The effect of the pre-trained teacher model.}}
  \label{fig:ablate_teacher}
\vspace{-10pt}

\end{figure}

\myheading{Study of Pretrained Teacher Model} results are shown in \cref{tab:ablation} (Settings E-F). We replace the pretrained depth diffusion model with two alternatives: Lotus \cite{he2024lotus} in setting (E), the same as \textbf{Ours}, and Marigold \cite{marigold} in setting (F). Setting (F) yields slightly better depth-accuracy metrics, but setting (E) demonstrates superior detail-accuracy, with a substantial margin of 50 in DBE completion. 
We provide qualitative results in~\cref{fig:ablate_teacher}.
Based on these findings, we select Lotus as the teacher model in our main experiments.

\myheading{Effect of Online vs. Offline Models} is examined in \cref{tab:ablation} (Settings G-H), where we explore the impact of online (our approach) and offline models on difference map calculation during training. We compare two settings: (G) using the static Lotus model and (H) using the EMA (exponential moving average) of the training model $\refinerema$. In setting (G), the offline model tends to overfit to the fixed differences between $\discri$ and $\genera$. In contrast, setting (H) dynamically learns from a difference map that progressively decreases as training advances. Empirically, setting (H) outperforms (G), so we adopt this approach in our main experiments.

%% file: sec/main_vis.tex
\begin{figure*}[!t]
    \vspace{0mm}
    \centering
    \includegraphics[width=0.85\linewidth]{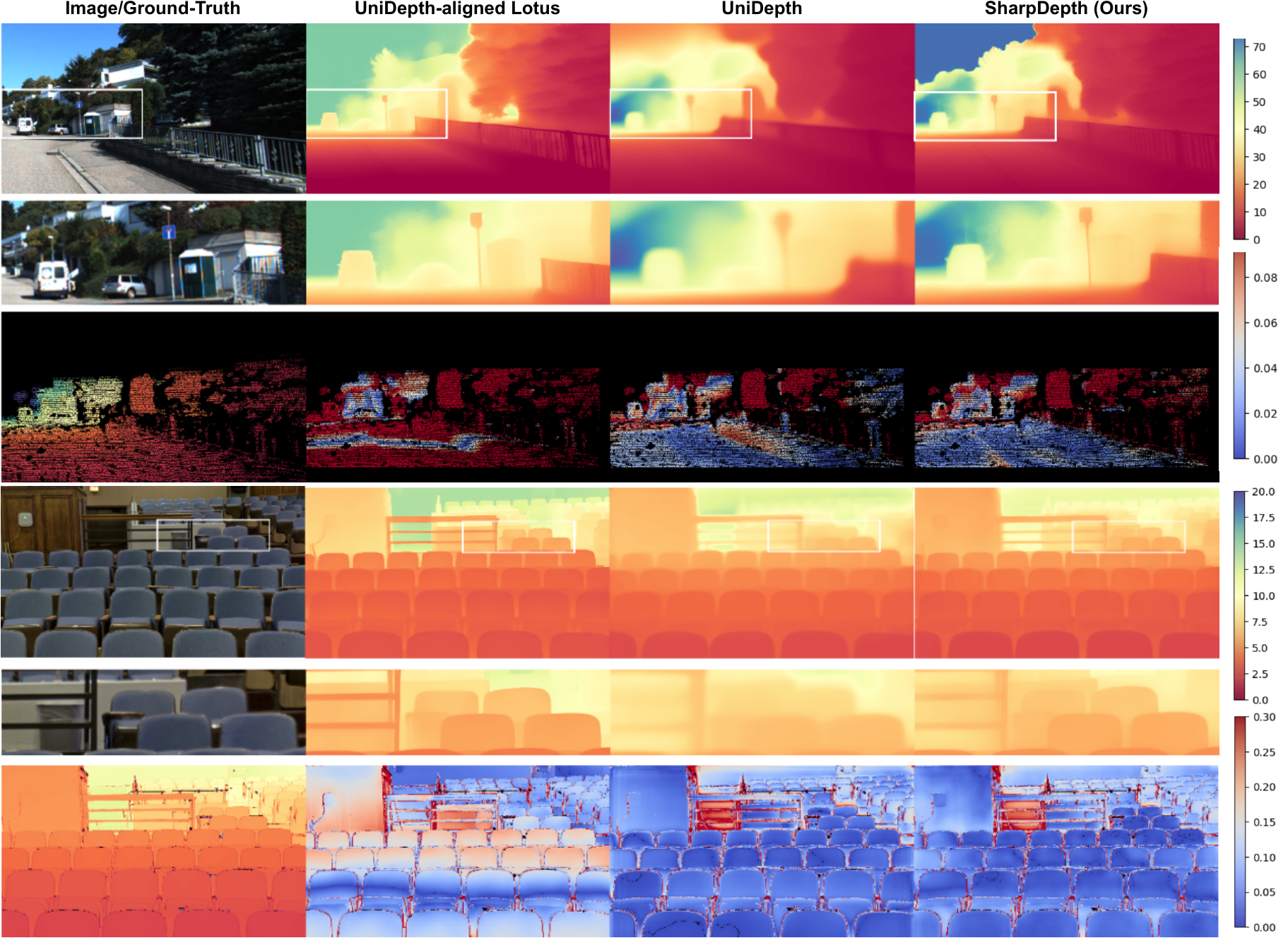}
    \caption{\textbf{Zero-shot qualitative results on unseen test samples of KITTI~\cite{geiger2013vision} and DIODE~\cite{vasiljevic2019diode} dataset.} Our method strikes a balance between depth accuracy and details. UniDepth lacks several details while UniDepth-aligned Lotus is less accurate.}
    \label{fig:results:main_vis}
\end{figure*}

%% file: sec/5_conclusion.tex
\section{Conclusion}
\label{sec:conclusion}
We proposed \Approach, a diffusion-based depth model that brings the metric precision of discriminative depth models into generative depth estimators. By balancing accuracy with detail, our method produces depth maps that are both metrically precise and visually refined, advancing high-quality zero-shot monocular depth estimation. Our evaluations show its strong performance across benchmarks, highlighting its potential for real-world applications requiring high-quality depth perception.


%% file: sec/X_suppl_arxiv.tex
In this supplementary material, we provide additional datasets details in~\cref{sec:implementation}. We then provide additional results in~\cref{sec:additional}. Finally, we demonstrate the effectiveness of our estimated depths in a downstream application implementing metric SLAM in~\cref{sec:appli}.

\section{Dataset Details}
\label{sec:implementation}
As described in the main paper, we train SharpDepth using approximately 1\% of the data from six real datasets and evaluate it on seven real datasets (for depth accuracy) and three synthetic datasets (for depth detail). 
This approach guarantees a diverse training dataset capable of encompassing various camera configurations. Details of the training and test sets are provided in \cref{tab:supp:datasets}.

\begin{table}[h]
    \centering
    \small
    \resizebox{\columnwidth}{!}{
    \begin{tabular}{clccc}
        \toprule
        & \textbf{Dataset} &  \textbf{Images} & \textbf{Scene} & \textbf{Acquisition}\\
        \midrule
        \multirow{6}{*}{\rotatebox[origin=c]{90}{\textbf{Training Set}}} 
        & Argoverse2~\cite{wilson2023argoverse} & 15k & Outdoor & LiDAR\\
        & Waymo~\cite{sun2020scalability} & 12k & Outdoor & LiDAR\\
        & PandaSet~\cite{xiao2021pandaset} & 12k & Outdoor & LiDAR\\
        & ARKit~\cite{baruch2021arkitscenes}  & 16k & Outdoor & LiDAR\\
        & ScanNet~\cite{dai2017scannet}  & 11k & Indoor & RGB-D\\
        & Taskonomy~\cite{zamir2018taskonomy} & 30k & Indoor & RGB-D\\
        
        \midrule
        \multirow{9}{*}{\rotatebox[origin=c]{90}{\textbf{Test Set}}}
        & KITTI~\cite{geiger2013vision} & 652 & Outdoor & LiDAR\\
        & NYU~\cite{silberman2012indoor} & 654 & Indoor & RGB-D\\
        & ETH3D~\cite{} & 454 & Outdoor & RGB-D\\
        & Diode~\cite{vasiljevic2019diode} & 325 & Indoor & LiDAR\\
        & Booster~\cite{ramirez2023booster}  & 456 & Indoor & RGB-D\\
        & NuScenes~\cite{caesar2020nuscenes}  & 1000 & Outdoor & LiDAR\\
        & IBims-1~\cite{ibims} & 100 & Indoor & RGB-D\\
        & Sintel~\cite{sintel}  & 1065 & Synthetic & - \\
        & UnrealStereo4K~\cite{unrealstereo} & 200 & Synthetic & -\\
        & Spring~\cite{mehl2023spring}  & 1016 & Synthetic & - \\
        
        \bottomrule
    \end{tabular}}
     \caption{\textbf{Datasets.} List of the training and test datasets along with their number of images, scene type, and acquisition method.} 
    \label{tab:supp:datasets}
\end{table}

\section{Additional Results}
\label{sec:additional}

\begin{figure*}
    \centering
    \includegraphics[width=\textwidth]{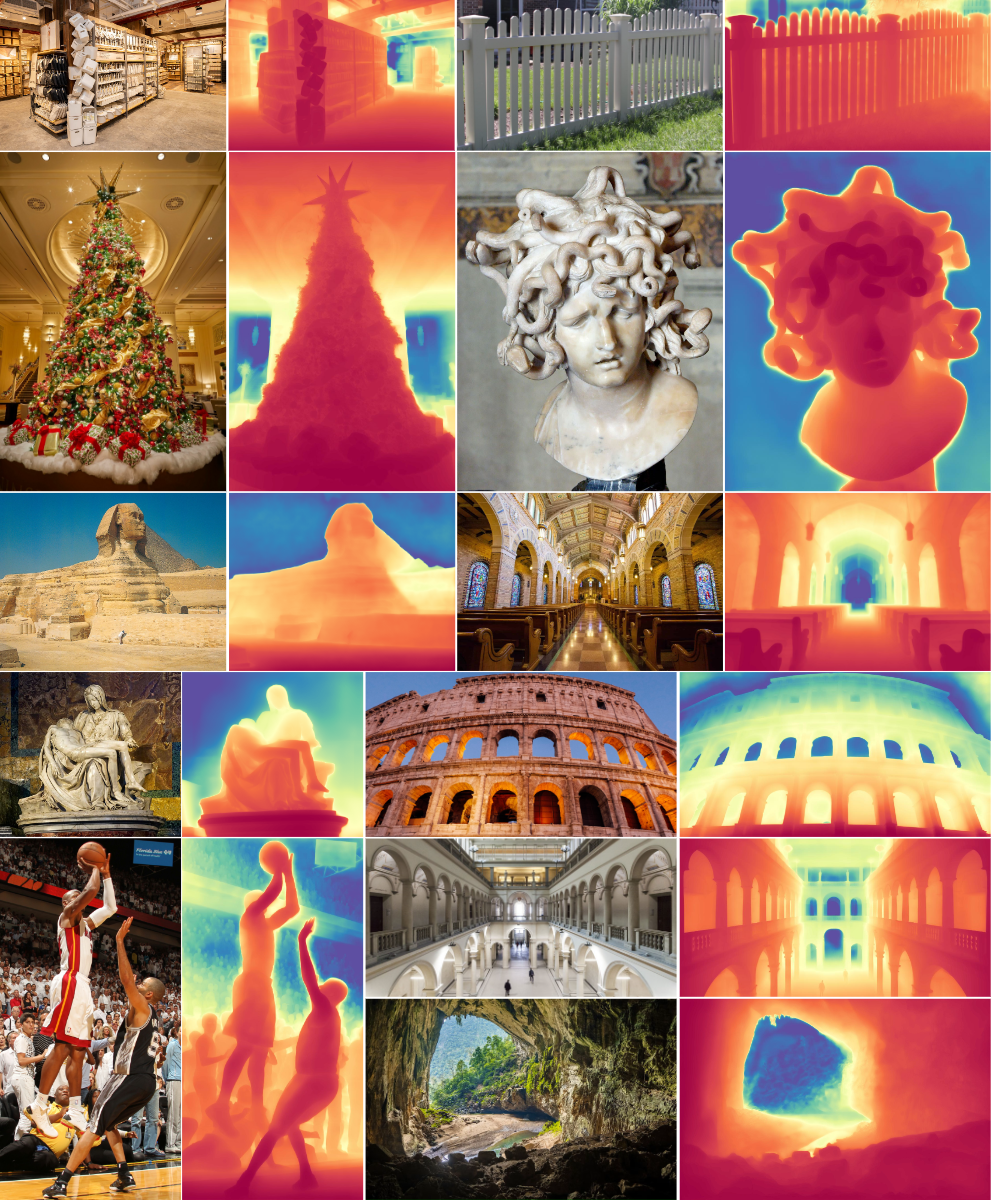}
    \caption{\textbf{In-the-wild depth estimation from Internet images}. Red indicates the close plane and blue means the far plane.}
    \label{fig:internet}
\end{figure*}

\begin{figure*}
    \centering
    \includegraphics[width=0.9\textwidth]{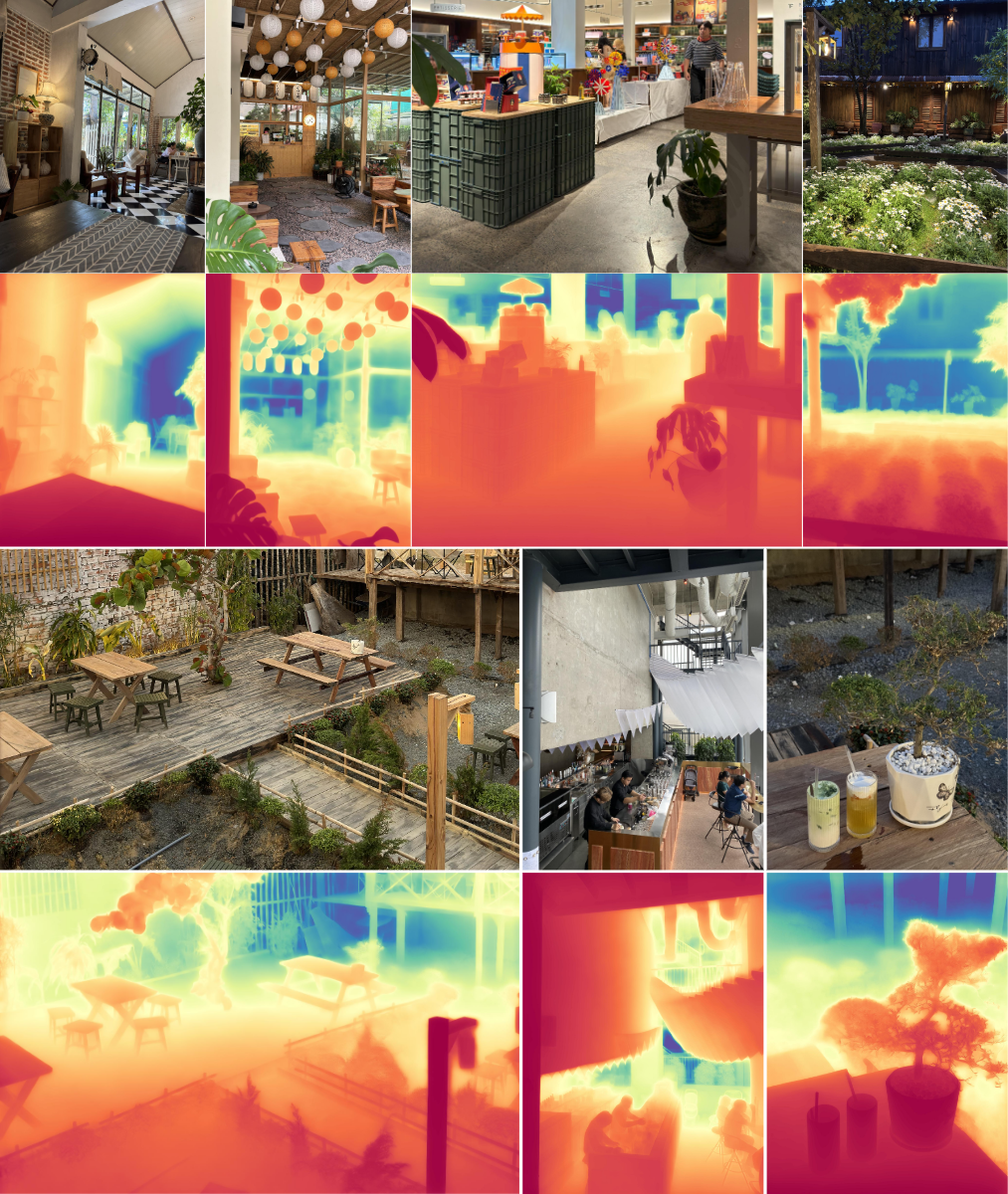}
    \caption{\textbf{In-the-wild depth estimation from images captured by a mobile phone}. Red indicates the close plane and blue means the far plane.}
    \label{fig:iphone}
\end{figure*}

\myheading{In-the-wild image samples.}
We evaluate the robustness of our method on a diverse set of ``in-the-wild'' images. Qualitative results for Internet-sourced images are shown in \cref{fig:internet}, while results from handheld mobile device captures are presented in \cref{fig:iphone}. Our approach consistently generates accurate metric depth maps, exhibiting improved depth discontinuities and overall structural coherence. Notably, it excels at capturing thin structures, comparable to affine-invariant diffusion depth models \cite{marigold, he2024lotus}, while preserving the precision of metric depth.

\begin{figure*}
    \centering
   \begin{subfigure}{0.8\linewidth}
        \caption{KITTI dataset}
        \vspace{-5pt}
        \includegraphics[width=\linewidth]{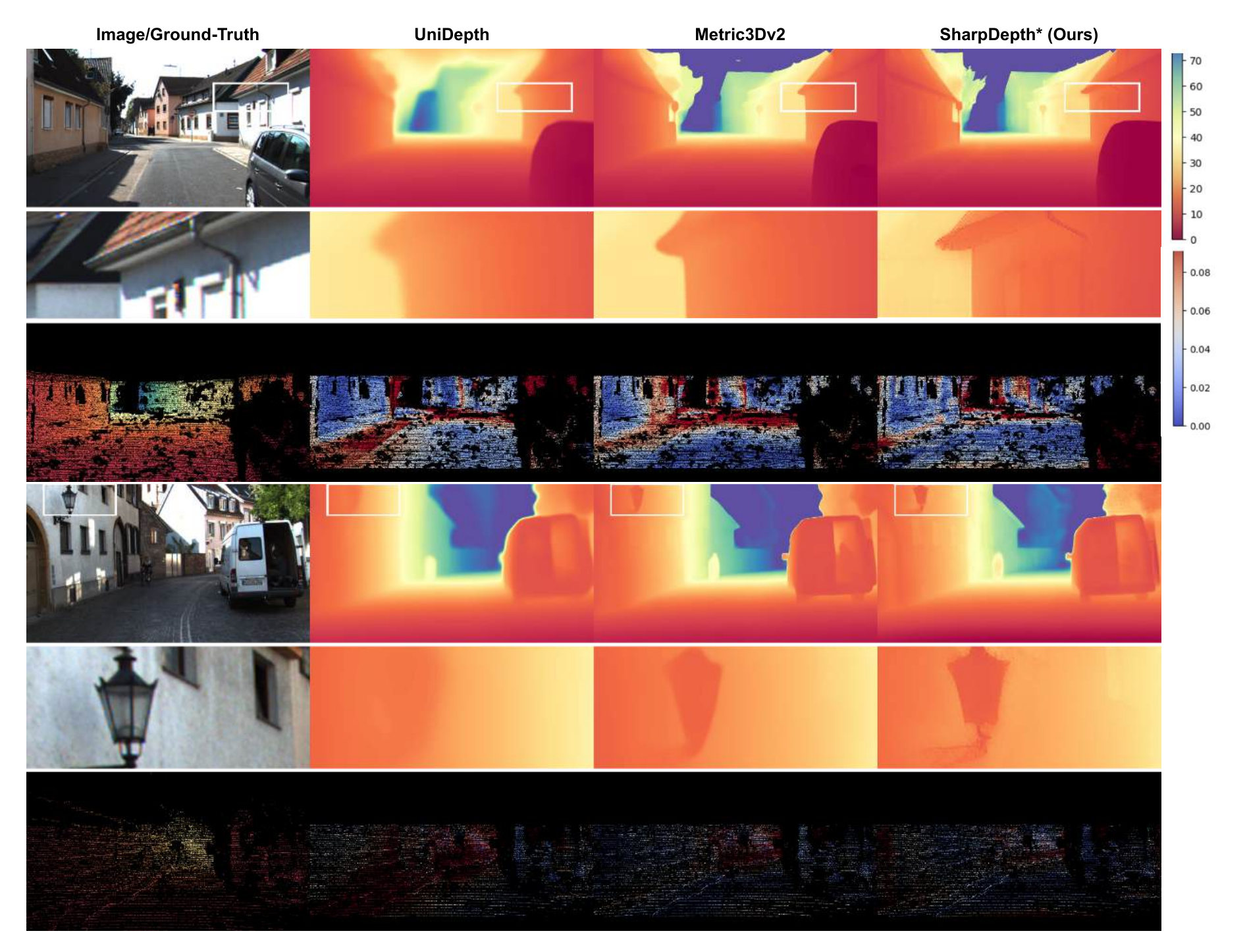}
        
   \end{subfigure}
    \vfill 
    \vspace{-3pt}
   \begin{subfigure}{0.75\linewidth}
        \caption{NYUv2 dataset}
        \vspace{-5pt}
        \includegraphics[width=\linewidth]{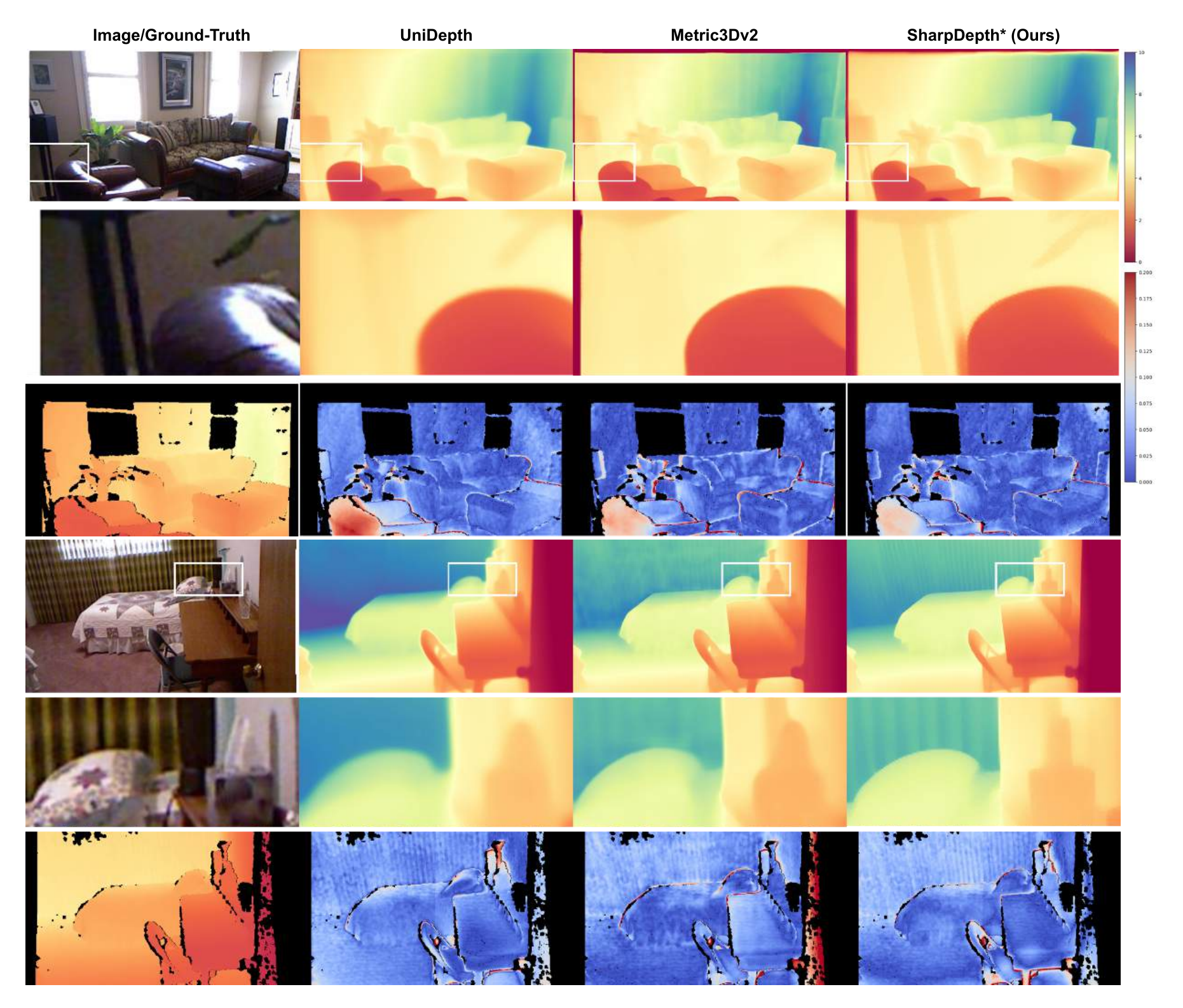}
        
    \end{subfigure}
    \vspace{-5pt}
    \caption{\textbf{Qualitative results on KITTI and NYUv2}. SharpDepth* denotes our method when using depth by Metric3Dv2 as input.}
    \label{fig:metric3dv2-quali}
    
\end{figure*}

\myheading{Generalization to another metric depth estimator.}
To show our method's generalization capabilities, we evaluate \Approach{} on Metric3Dv2~\cite{hu2024metric3d}, a recent versatile metric depth estimation model. In this experiment, we leverage our previously trained model on UniDepth, and no further retraining is performed. We directly apply our model trained on UniDepth to Metric3Dv2 depths during test time. We provide the qualitative results in \cref{fig:metric3dv2-quali}. As can be seen, our method generalizes well to metric depths produced by Metric3Dv2.

\begin{figure*}[t]
    \centering
    \includegraphics[width=0.85\linewidth]{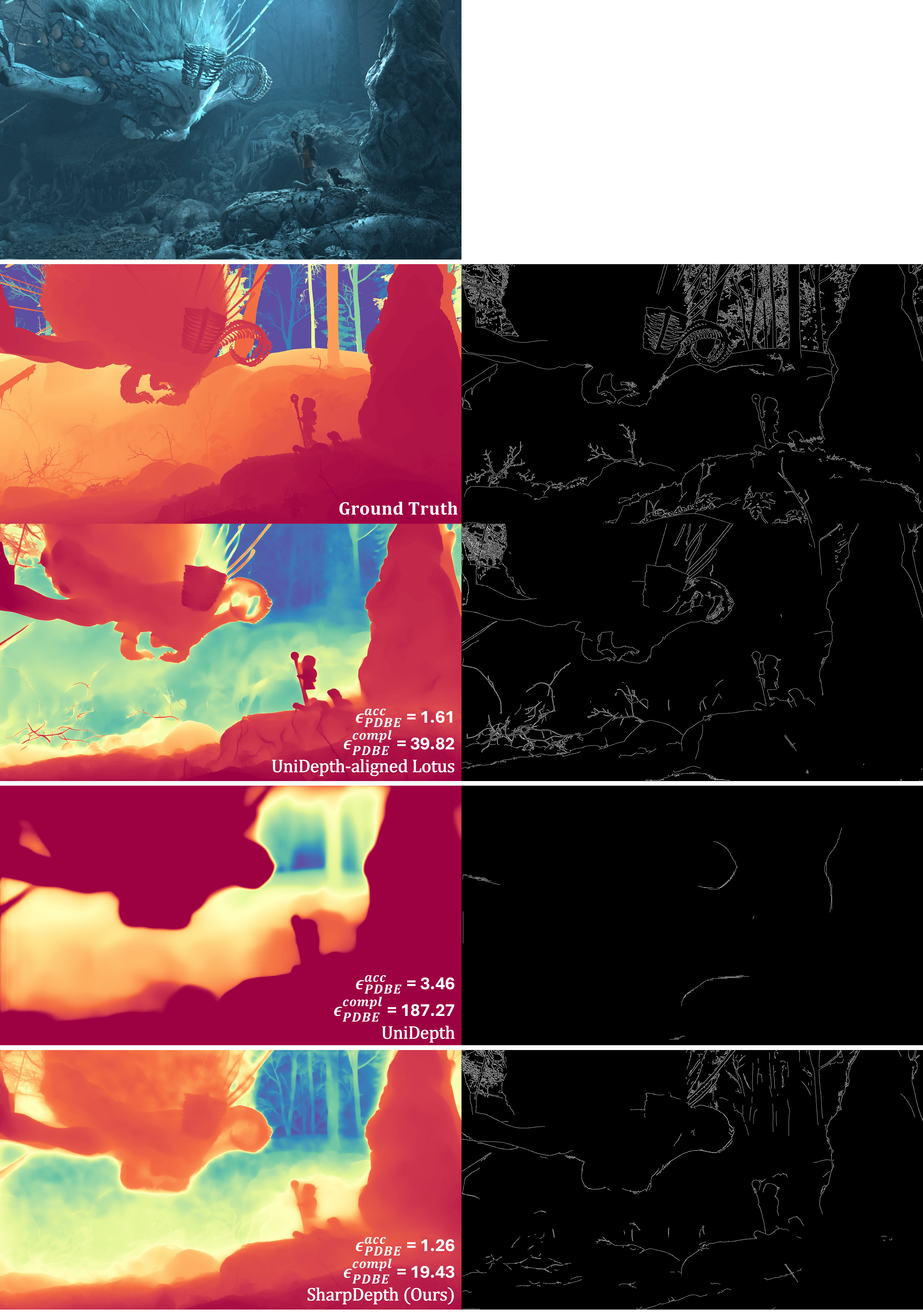}
    \caption{\textbf{Illustration of the depth boundary metrics on the Spring dataset}. We show the depth maps and extracted boundaries for each prediction. Compared to UniDepth, our method extracts more edges due to better depth discontinuities. Compared to Lotus, our method can capture more precise edges, due to the global prior from pre-trained UniDepth.}
    \label{fig:metric-illus-0}
\end{figure*}

\begin{figure*}[t]
    \centering
    \includegraphics[width=0.85\linewidth]{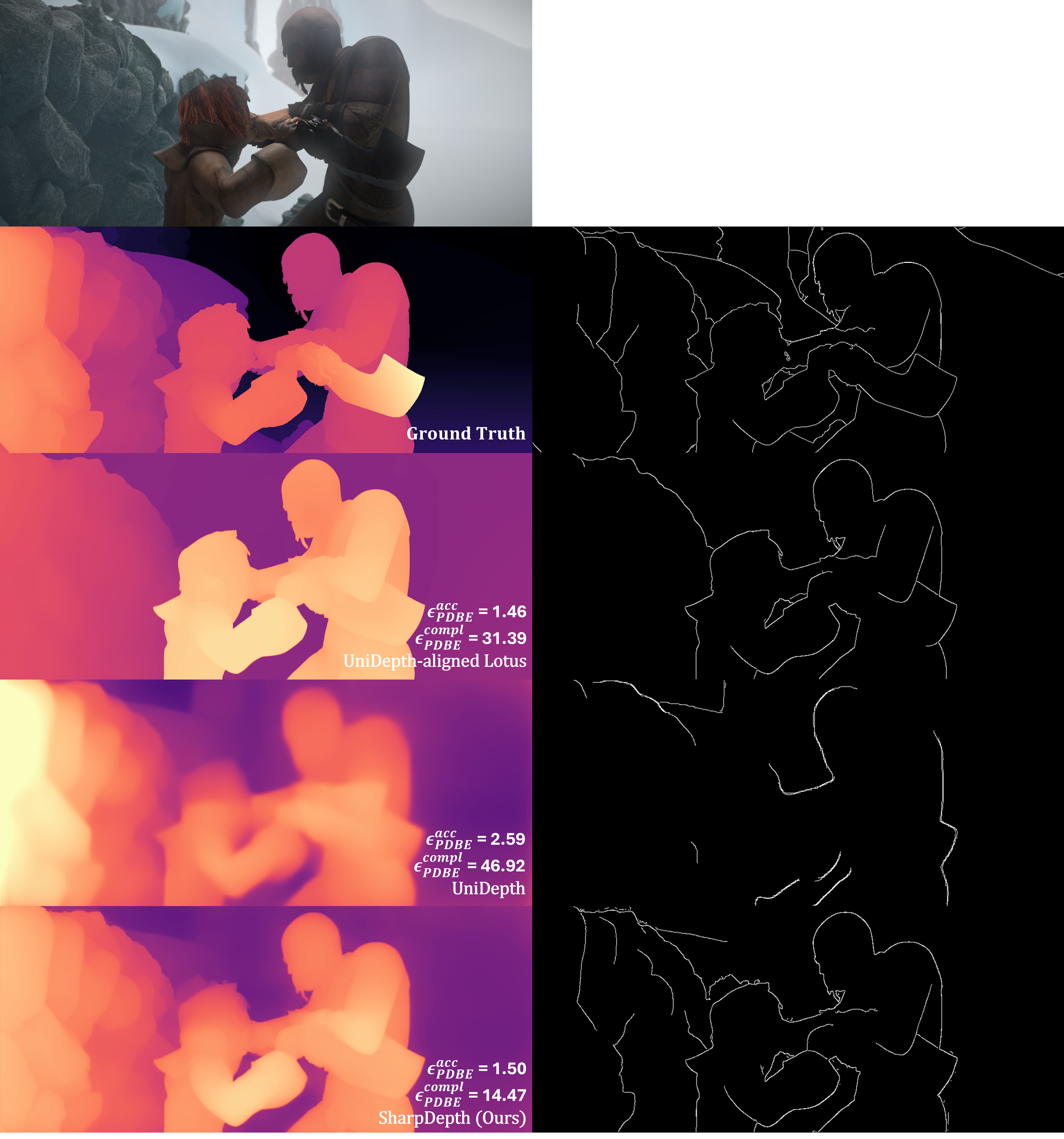}
    \caption{\textbf{Illustration of the depth boundary on the Sintel dataset}. We show the depth maps and extracted boundaries for each prediction.}
    \label{fig:metric-illus-1}
\end{figure*}

\myheading{More depth metrics on test datasets.}
We provide an extended version of zero-shot metric accuracy on 6 zero-shot datasets in \cref{tab:supp:full_results}. We report absolute mean relative error (A.Rel), root mean square error (RMSE), scale-invariant error in log scale ($\text{SI}_{\text{log}})$ and the percentage of inlier pixel ($\mathrm{\delta}_{1}$).
As shown in \cref{tab:supp:full_results}, \Approach~achieves competitive metric accuracy compared to UniDepth and other metric depth models. Moreover, our method consistently outperforms other metric refinement techniques, such as PatchRefiner. This highlights the effectiveness of our approach in enhancing high-frequency details in depth maps while maintaining robust zero-shot performance.
We further report the Pseudo Depth Boundary Error (PDBE), including the accuracy $\epsilon^{\text{acc}}_{\text{PDBE}}$ and completion $\epsilon^{\text{compl}}_{\text{PDBE}}$, along with visual samples in \cref{fig:metric-illus-0} and \cref{fig:metric-illus-1}. The results demonstrate that both the accuracy and completion rates effectively capture the boundary details of the depth map.

\begin{figure*}
    \centering
    \vspace{-10pt}
   \begin{subfigure}{0.83\textwidth}
        \caption{KITTI dataset}
        \vspace{-5pt}
        \includegraphics[width=\linewidth]{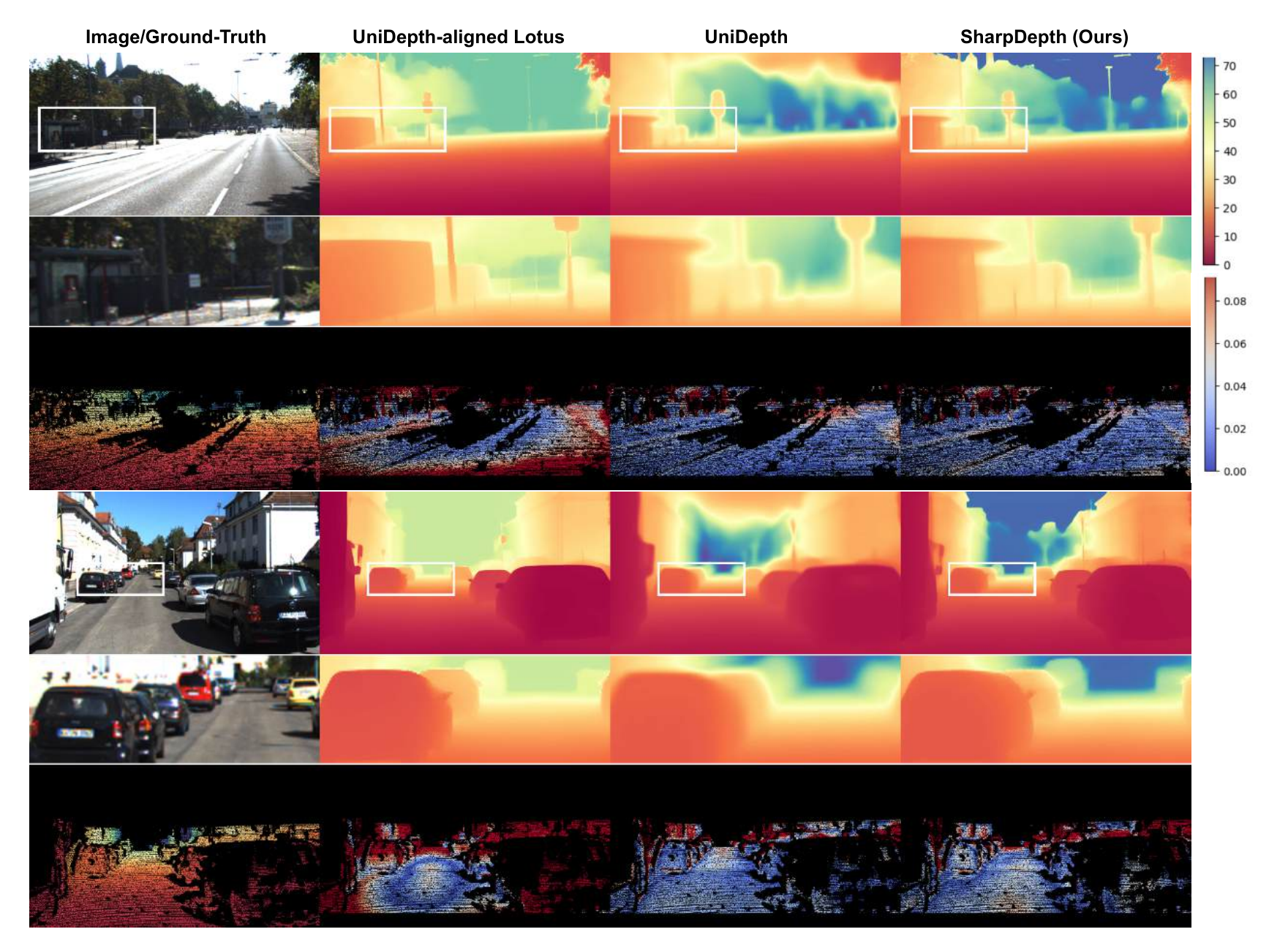}
        
   \end{subfigure}
    \vfill 
    \vspace{-10pt}
   \begin{subfigure}{0.83\textwidth}
        \caption{NYUv2 dataset}
        \vspace{-5pt}
        \includegraphics[width=\linewidth]{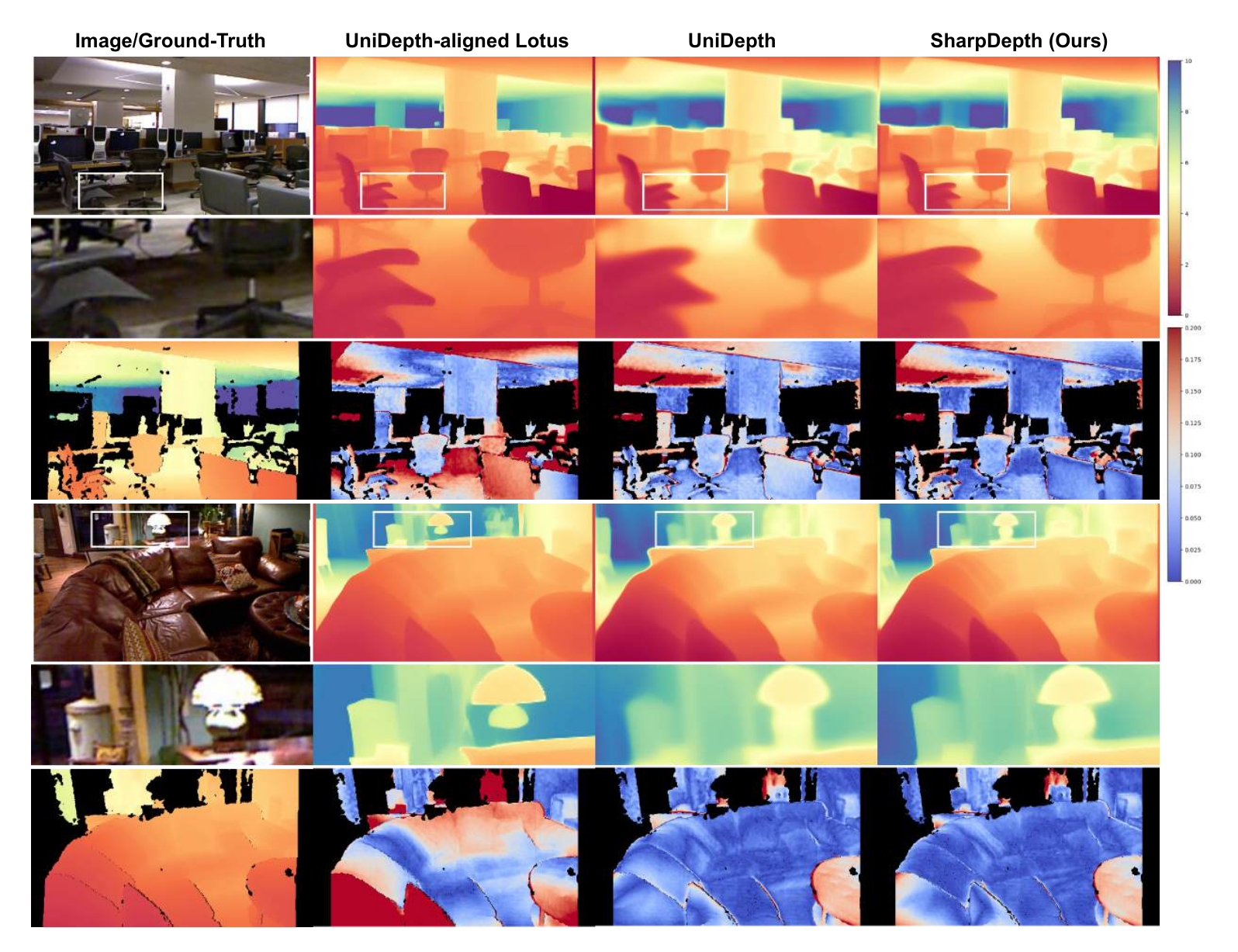}
        
    \end{subfigure}
    \caption{\textbf{Qualitative comparisons on different datasets} (1/3). }
    \label{fig:cont1}

\end{figure*}

\begin{figure*}
    \centering
   \begin{subfigure}{0.8\linewidth}
        \caption{ETH3D dataset}
        \vspace{-5pt}
        \includegraphics[width=\linewidth]{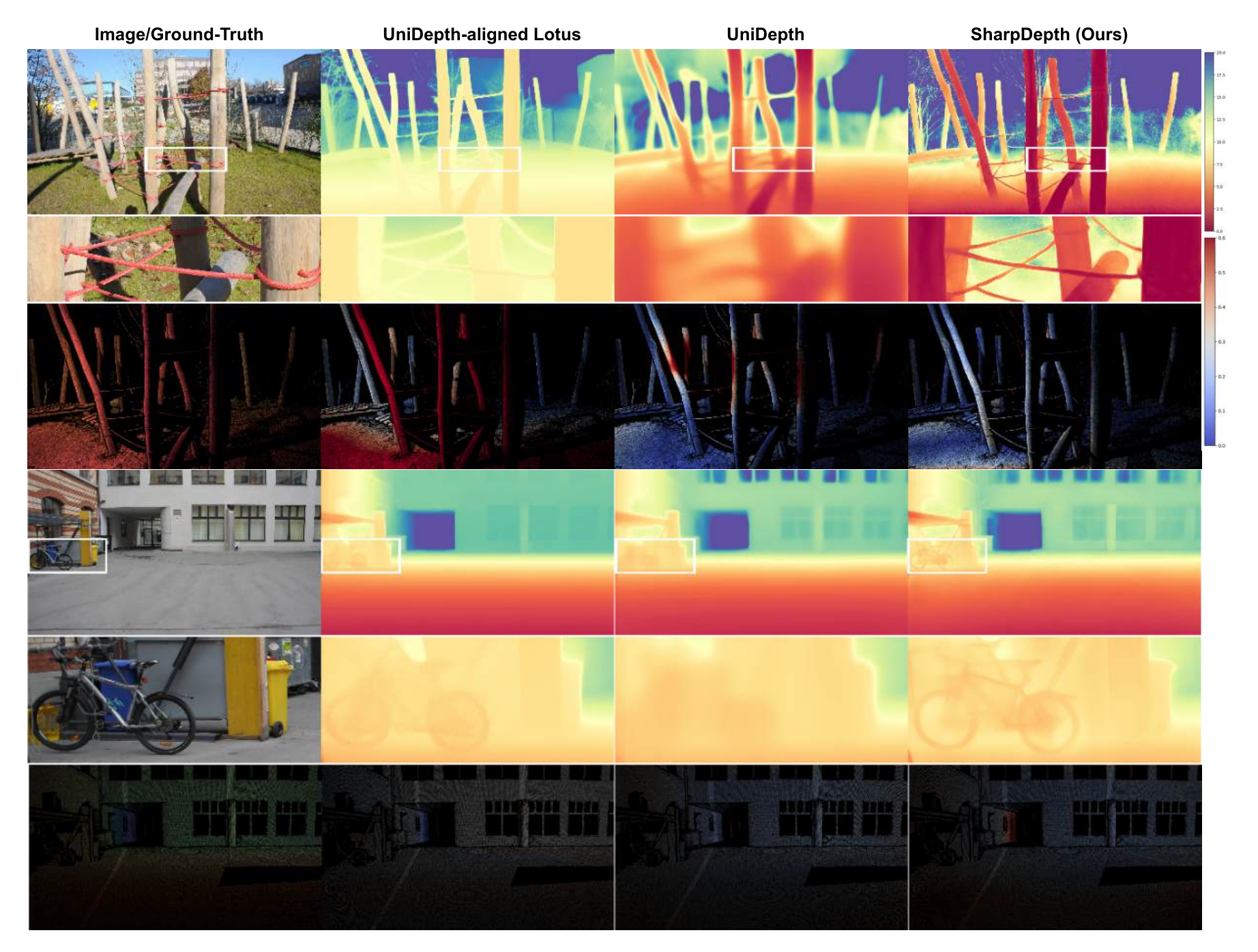}
        
    \end{subfigure}
    \vfill
    \vspace{-10pt}
   \begin{subfigure}{0.8\linewidth}
        \caption{Diode dataset}
        \vspace{-5pt}
        \includegraphics[width=\linewidth]{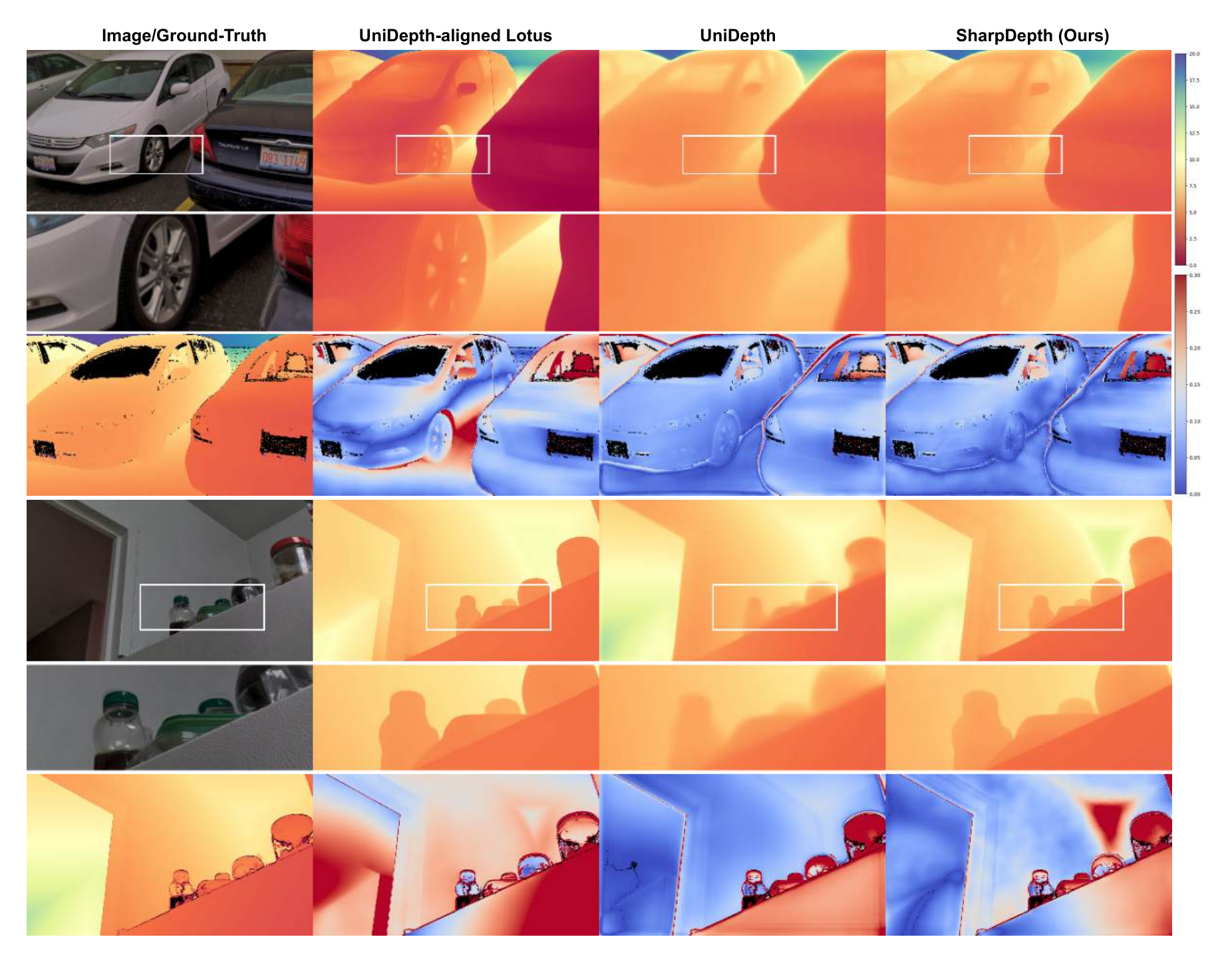}
        
    \end{subfigure}
    \vfill
    \caption{\textbf{Qualitative comparisons on different datasets} (2/3). }
    \label{fig:cont2}
    
\end{figure*}

\begin{figure*}
    \centering
   \begin{subfigure}{0.8\linewidth}
        \caption{Booster dataset}
        \vspace{-5pt}
        \includegraphics[width=\linewidth]{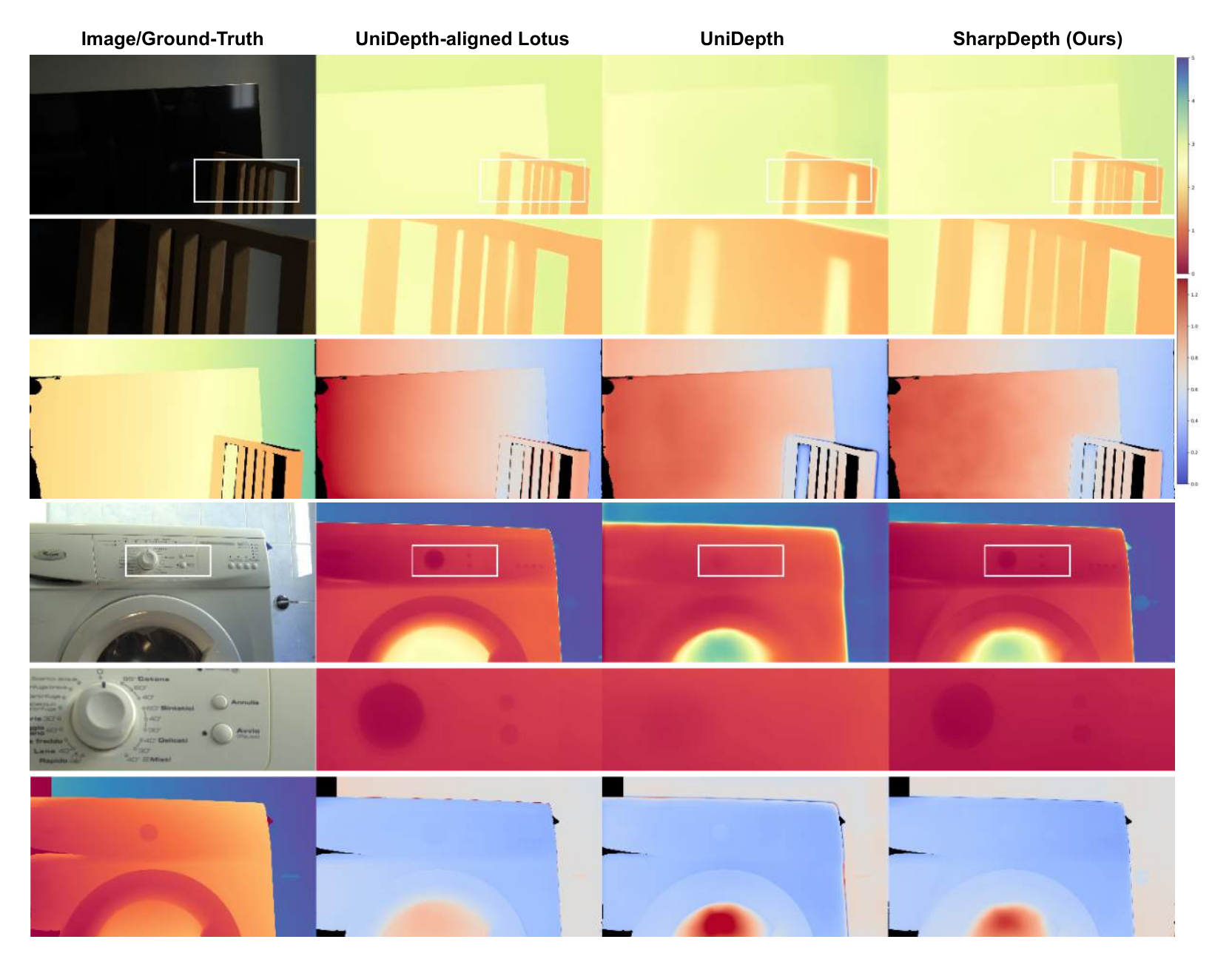}
        
    \end{subfigure}
    \vfill
   \begin{subfigure}{0.8\linewidth}
        \caption{nuScenes dataset}
        \vspace{-5pt}
        \includegraphics[width=\linewidth]{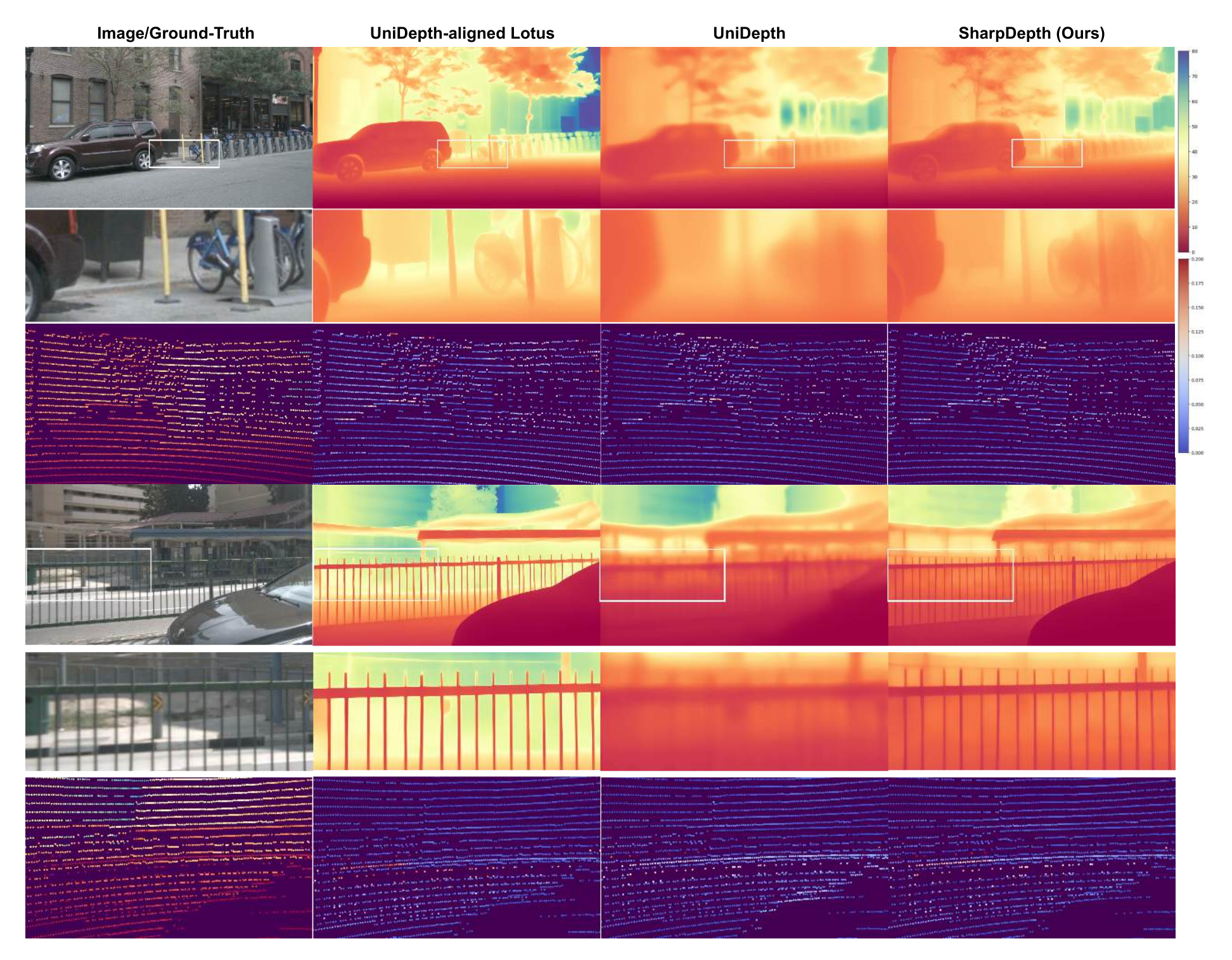}
        
    \end{subfigure}
    \caption{\textbf{Qualitative comparisons on different datasets} (3/3). }
    \label{fig:cont3}
\end{figure*}

\myheading{More visual results.}
We present additional qualitative results on our test datasets in \cref{fig:cont1}, \cref{fig:cont2}, and \cref{fig:cont3}. These figures show predictions from UniDepth, UniDepth-aligned Lotus, and our method. As observed, our approach generates depth maps with more detailed representations of fine structures.

\begin{table}[t]
    \centering
    \resizebox{\columnwidth}{!}{
    \begin{tabular}{lccc}
        \toprule
        \textbf{Method} & \textbf{PSNR}$\uparrow$ &\textbf{SSIM}$\uparrow$ & \textbf{LPIPS}$\downarrow$  \\
        \toprule
        MonoGS + UniDepth & 18.472 & 0.718 & 0.305 \\
        MonoGS + Ours & \textbf{18.857} &\textbf{ 0.735} & \textbf{0.289}  \\
        \bottomrule
    \end{tabular}}
    \caption{Performance of~\Approach{} on the fr1/desk sequence of TUM RGB-D dataset \cite{TumRGBD}. 
    }
    \label{tab:supp:slam}
    \vspace{-5pt}
\end{table}

\section{Applications}
\label{sec:appli}

In this section, we demonstrate that our predicted sharper depth maps can significantly benefit downstream 3D reconstruction tasks, such as Visual SLAM~\cite{guizilini2022full} and Volumetric TSDF Fusion~\cite{zeng20163dmatch}. By providing more detailed and accurate depth information, our method enhances the quality and reliability of these reconstruction pipelines.

\subsection{Visual SLAM}
\label{sec:appli_slam}

\begin{figure*}[t]
  \centering
  \includegraphics[width=0.9\linewidth]{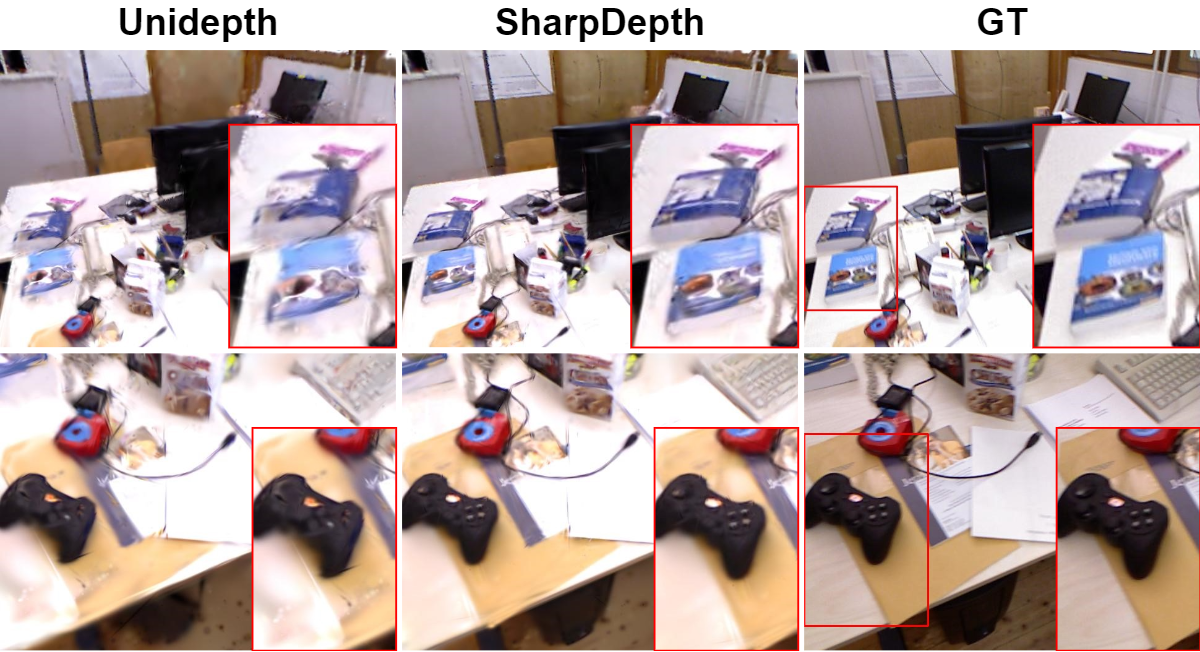}
   \caption{\textbf{Rendering comparison on TUM fr1/desk sequence}. For each method, we show the novel view rendering. Compared to UniDepth (leftmost column), using SharpDepth (middle column) can result in finer details of objects, such as the books in the first row and the game console in the second row.}
  \label{fig:slam}
\end{figure*}

Dense visual SLAM focuses on reconstructing detailed 3D maps, which are crucial for applications in AR and robotics. In this work, we demonstrate that high-frequency depth maps can significantly improve the performance of SLAM methods in reconstructing the scene.
We conduct experiments using a Gaussian Splatting-based SLAM method, i.e., MonoGS \cite{MonoGS}, on the fr1/desk sequence of TUM RGBD dataset \cite{TumRGBD}, using the depth maps from UniDepth and \Approach~as inputs to the system. Quantitative results are provided in \cref{tab:supp:slam}, where our method consistently outperforms UniDepth in terms of photometric errors, showcasing its potential to enhance SLAM performance.
Additionally, we present qualitative results in \cref{fig:slam}. As shown, our method better captures the underlying geometry of the scene, leading to improved novel view renderings.

\subsection{Volumetric TSDF Fusion}
\label{sec:appli_tsdf}

Existing 3D reconstruction pipelines rely on multiple pairs of RGB-D inputs that are multi-view consistent. To achieve high-quality point clouds, it is crucial to have accurate metric depth predictions with sharp details. In this section, we demonstrate that our predicted depth maps can be used with TSDF Fusion~\cite{zeng20163dmatch}, to further enhance their reconstruction quality. 

\begin{figure*}[t]
  \centering
  \includegraphics[width=0.9\linewidth]{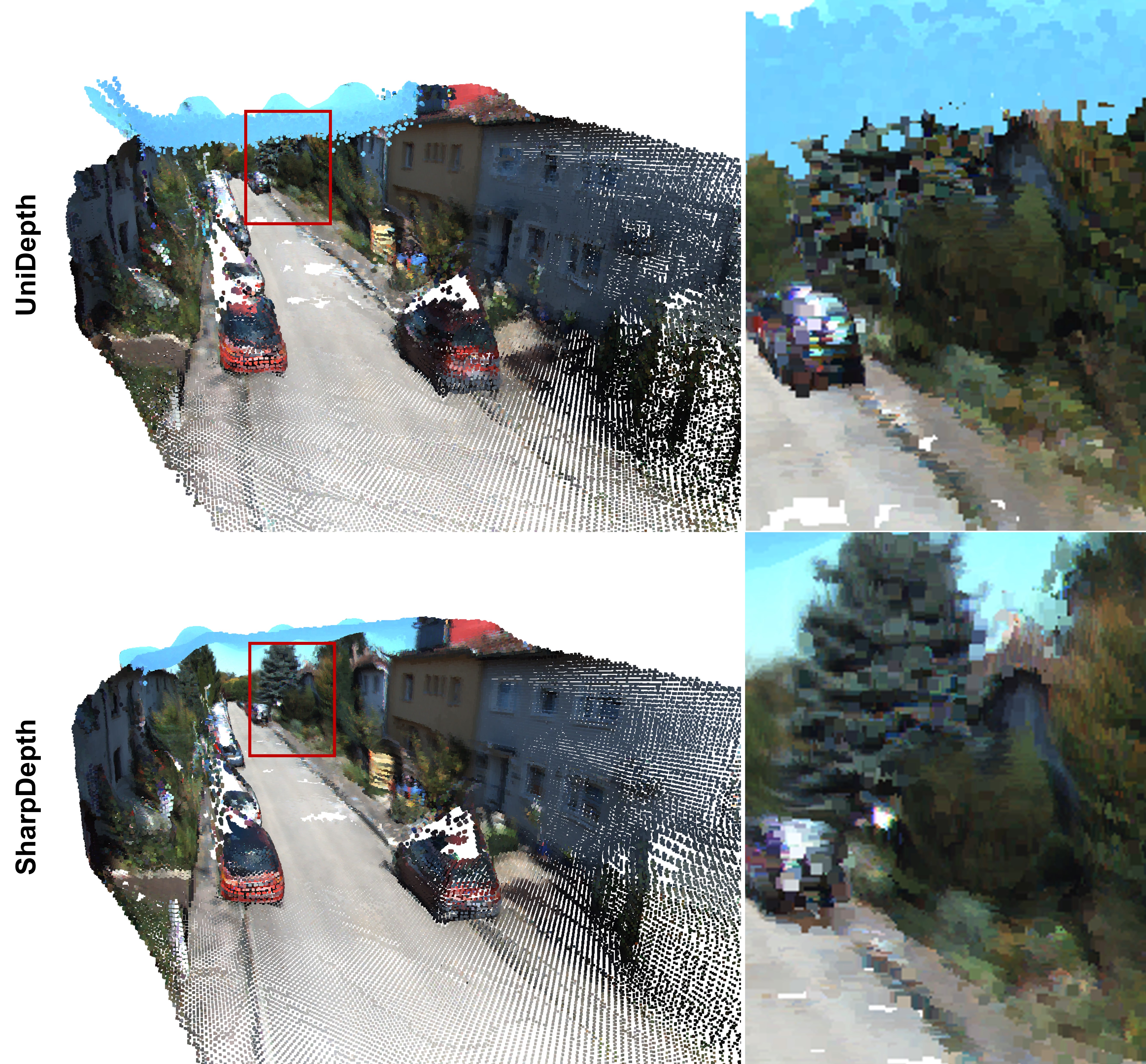}
   \caption{\textbf{Multi-view scene reconstruction on KITTI dataset}. We predict depth maps using UniDepth and SharpDepth for each frame and use TSDF-Fusion to generate the point cloud. SharpDepth's point cloud achieves less shape distortion in vehicles. }
  \label{fig:tsdf}
\end{figure*}

As can be seen in~\cref{fig:tsdf}, SharpDepth can render less distorted point clouds compared to those produced by the UniDepth~\cite{piccinelli2024unidepth} approach.

\begin{table*}[]
    \centering
   \small
    \setlength{\cmidrulewidth}{0.5pt}
    \vspace{-10pt}
    \begin{tabular}{llcccc}
        \toprule
       \textbf{Dataset} & \textbf{Method} & $\mathbf{\mathrm{A.Rel}\downarrow}$ & $\mathbf{\mathrm{RMSE}\downarrow}$ & $\mathrm{SI}_{\log}\downarrow$ & $\mathrm{\delta}_{1}\uparrow$\\
        \toprule
      \multirow{8}{*}{KITTI} & Marigold~\cite{marigold} & 0.095 & 3.221 & 13.240 & 92.284\\
        & Lotus~\cite{he2024lotus}  & 0.113 & 3.538 & 18.383 & 87.703 \\
         \cmidrule(lr){2-6}
        & UniDepth~\cite{piccinelli2024unidepth} & \cellcolor{tabfirst!170} 0.051 & \cellcolor{tabfirst!170} 2.236 & \cellcolor{tabfirst!170} 7.078 & \cellcolor{tabfirst!170} 97.921 \\
        & ZoeDepth~\cite{zoedepth} &\cellcolor{gray!50} 0.057 & \cellcolor{gray!50}2.390 & \cellcolor{gray!50}7.470 & \cellcolor{gray!50}96.500\\
        & Metric3Dv2~\cite{hu2024metric3d} & \cellcolor{tabsecond!170} 0.053 & \cellcolor{tabthird!170} 2.481 & \cellcolor{tabsecond!170} 7.449 & \cellcolor{tabsecond!170} 97.589\\
        & PatchRefiner~\cite{li2025patchrefiner} & 0.158 & 6.043 & 13.061 & 79.245\\
        & UniDepth-aligned Lotus & 0.130 & 3.935 &  16.077 & 83.633 \\
        \cmidrule(lr){2-6}
        & ~\Approach~(Ours) & \cellcolor{tabthird!170} 0.059 & \cellcolor{tabsecond!170} 2.374 & \cellcolor{tabthird!170} 8.100 & \cellcolor{tabthird!170} 97.315 \\
        \midrule
        
      \multirow{8}{*}{NYUv2}  & Marigold~\cite{marigold} & 0.055 & 0.224 & 8.114 & 96.384\\
        & Lotus~\cite{he2024lotus} & 0.054 & 0.222 & 7.993 & 96.612\\
        \cmidrule(lr){2-6}
        & UniDepth~\cite{piccinelli2024unidepth} & \cellcolor{tabfirst!170} 0.055 & \cellcolor{tabfirst!170} 0.200 & \cellcolor{tabfirst!170} 5.367 & \cellcolor{tabfirst!170} 98.417\\
        & ZoeDepth~\cite{zoedepth} &\cellcolor{gray!50} 0.077 & \cellcolor{gray!50}0.278 & \cellcolor{gray!50}7.190 & \cellcolor{gray!50}95.200\\
        & Metric3Dv2~\cite{hu2024metric3d} & \cellcolor{tabthird!170} 0.066 & \cellcolor{tabthird!170} 0.254 & \cellcolor{tabthird!170} 7.498 &  \cellcolor{tabsecond!170}  97.391\\
        & PatchRefiner~\cite{li2025patchrefiner} & 2.482 & 5.900 & 19.089 & 1.000\\
        & UniDepth-aligned Lotus & 0.087 & 0.281 & 8.916 & 93.921 \\
        \cmidrule(lr){2-6}
        & ~\Approach~(Ours) & \cellcolor{tabsecond!170} 0.064 & \cellcolor{tabsecond!170} 0.228 & \cellcolor{tabsecond!170}  6.179 & \cellcolor{tabthird!170} 96.949\\
        \midrule
      \multirow{8}{*}{ETH3D}  & Marigold~\cite{marigold} & 0.064 & 0.616 & 9.217 & 95.956 \\
        & Lotus~\cite{he2024lotus} & 0.062 & 0.581 & 9.266 & 96.001 \\
        \cmidrule(lr){2-6}
        & UniDepth~\cite{piccinelli2024unidepth} & \cellcolor{tabsecond!170} 0.456 & \cellcolor{tabsecond!170} 3.008 & \cellcolor{tabsecond!170} 7.728 &\cellcolor{tabthird!170}  25.308\\
        & ZoeDepth~\cite{zoedepth} & 0.567 & 3.272 & 13.015 & \cellcolor{tabsecond!170} 34.210\\
        & Metric3Dv2~\cite{hu2024metric3d} & \cellcolor{tabfirst!170} 0.138 & \cellcolor{tabfirst!170} 0.903 & \cellcolor{tabfirst!170} 6.081 & \cellcolor{tabfirst!170} 82.420\\
        & PatchRefiner~\cite{li2025patchrefiner} & 1.781 & 8.830 & \cellcolor{tabthird!170} 11.715 & 4.974\\
        & UniDepth-aligned Lotus & 0.493 & 3.267 & 13.092 & 20.347\\
        \cmidrule(lr){2-6}
        & ~\Approach~(Ours) & \cellcolor{tabthird!170} 0.474 & \cellcolor{tabthird!170} 3.092 & 12.119 & 22.606\\
        \midrule
        
      \multirow{8}{*}{Diode}  & Marigold~\cite{marigold} & 0.307 & 3.755 & 29.230 & 76.685\\
        & Lotus~\cite{he2024lotus} & 0.330 & 3.877 & 30.999 & 73.751\\
        \cmidrule(lr){2-6}
        & UniDepth~\cite{piccinelli2024unidepth} & \cellcolor{tabsecond!170} 0.265 & \cellcolor{tabsecond!170} 4.216 & \cellcolor{tabsecond!170} 23.370 & \cellcolor{tabsecond!170} 66.031\\
        & ZoeDepth~\cite{zoedepth} & 0.484 & 6.637 & 29.374 & 30.195 \\
        & Metric3Dv2~\cite{hu2024metric3d} & \cellcolor{tabfirst!170} 0.158 & \cellcolor{tabfirst!170} 2.552 & \cellcolor{tabfirst!170} 19.455 & \cellcolor{tabfirst!170} 88.765\\
        & PatchRefiner~\cite{li2025patchrefiner} & 1.264 & 7.064 & 29.563 & 25.031\\
        & UniDepth-aligned Lotus & 0.357 & 5.321 & 30.671 & 55.876\\
        \cmidrule(lr){2-6}
        & ~\Approach~(Ours) & \cellcolor{tabthird!170} 0.297 & \cellcolor{tabthird!170} 4.644 & \cellcolor{tabthird!170} 25.340 & \cellcolor{tabthird!170} 61.486\\
        \midrule
        
      \multirow{8}{*}{Booster}  & Marigold~\cite{marigold} & 0.049 & 0.074 & 6.392 & 97.384\\
        & Lotus~\cite{he2024lotus} & 0.041 & 0.063 & 5.333 & 98.779\\
        \cmidrule(lr){2-6}
        & UniDepth~\cite{piccinelli2024unidepth} &  \cellcolor{tabsecond!170}0.492 &  \cellcolor{tabthird!170} 0.532 & \cellcolor{tabthird!170} 7.686 & \cellcolor{tabfirst!170} 28.041 \\
        & ZoeDepth~\cite{zoedepth} & 0.642 & 0.674 & 10.563 & 20.855\\
        & Metric3Dv2~\cite{hu2024metric3d} & 0.668 & 0.720 & 5.795 &  15.490\\
        & PatchRefiner~\cite{li2025patchrefiner} & 5.551 & 5.994 & 18.136 & 1.000 \\
        & UniDepth-aligned Lotus & \cellcolor{tabthird!170} 0.494 & \cellcolor{tabfirst!170} 0.519 & \cellcolor{tabfirst!170} 6.382 & \cellcolor{tabthird!170} 26.429\\
        \cmidrule(lr){2-6}
        & ~\Approach~(Ours) & \cellcolor{tabfirst!170} 0.491 & \cellcolor{tabsecond!170} 0.528 & \cellcolor{tabsecond!170} 7.089 & \cellcolor{tabsecond!170} 27.717 \\
        \midrule
      \multirow{8}{*}{nuScenes}  & Marigold~\cite{marigold} & 0.267 & 6.158 & 35.628 & 65.881\\
        & Lotus~\cite{he2024lotus} & 0.363 & 7.263 & 49.047 & 50.911\\
        \cmidrule(lr){2-6}
        & UniDepth~\cite{piccinelli2024unidepth} & \cellcolor{tabfirst!170} 0.144 & \cellcolor{tabfirst!170} 4.771 & \cellcolor{tabfirst!170} 21.959 &  \cellcolor{tabsecond!170} 83.861 \\
        & ZoeDepth~\cite{zoedepth} & 0.587 & 8.155 & 33.076 & 21.838  \\
        & Metric3Dv2~\cite{hu2024metric3d} & \cellcolor{tabthird!170} 0.199 & \cellcolor{tabthird!170} \cellcolor{tabthird!170} 7.371 & \cellcolor{tabthird!170} 28.267 & \cellcolor{tabfirst!170} 84.215\\
        & PatchRefiner~\cite{li2025patchrefiner} & 0.582 & 10.589 & 30.193 & 31.726\\
        & UniDepth-aligned Lotus & 0.432 & 7.850 & 49.524 & 41.243 \\
        \cmidrule(lr){2-6}
        & ~\Approach~(Ours) & \cellcolor{tabsecond!170} 0.184 & \cellcolor{tabsecond!170} 5.208 & \cellcolor{tabsecond!170} 25.584 & \cellcolor{tabthird!170} 78.479\\

        \bottomrule

    \end{tabular}
     \caption{\textbf{Detailed results on different datasets.} We ranked methods that do not require GT alignment as \colorbox{tabfirst!170}{best}, \colorbox{tabsecond!170}{second-best}, and \colorbox{tabthird!170}{third-best}. \colorbox{gray!50}{Gray} indicates the method that has been trained on the training set.}
    \label{tab:supp:full_results}
    \vspace{-5pt}
\end{table*}

